# Leo: Lagrange Elementary Optimization

Aso M. Aladdin [1] *, Tarik A. Rashid[2]

[1]Department of Computer Science, College of Science, Charmo University, 46023, Chamchamal/Sulaimani, lraq.
aso.aladdin@charmouniversity.org
[2]Computer Science and Engineering Department, University of Kurdistan Hewler, Erbil 44001, Iraq.
tarik.ahmed@ukh.edu.krd

**Abstract :**Global optimization problems are frequently solved using the practical and efficient method of evolutionary sophistication. But as the original problem becomes more complex, so does its efficacy and expandability. Thus, the purpose of this research is to introduce the Lagrange Elementary Optimization (Leo) as an evolutionary method, which is self-adaptive inspired by the remarkable accuracy of vaccinations using the albumin quotient of human blood. They develop intelligent agents using their fitness function value after gene crossing. These genes direct the search agents during both exploration and exploitation. The main objective of the Leo algorithm is presented in this paper along with the inspiration and motivation for the concept. To demonstrate its precision, the proposed algorithm is validated against a variety of test functions, including 19 traditional benchmark functions and the CECC06 2019 test functions. The results of Leo for 19 classic benchmark test functions are evaluated against DA, PSO, and GA separately, and then two other recent algorithms such as FDO and LPB are also included in the evaluation. In addition, the Leo is tested by ten functions on CECC06 2019 with DA, WOA, SSA, FDO, LPB, and FOX algorithms distinctly. The cumulative outcomes demonstrate Leo's capacity to increase the starting population and move toward the global optimum. Different standard measurements are used to verify and prove the stability of Leo in both the exploration and exploitation phases. Moreover, Statistical analysis supports the findings results of the proposed research. Finally, novel applications in the real world are introduced to demonstrate the practicality of Leo.

## 1. Introduction

Numerous scientific and engineering disciplines emphasize the importance of stochastic parabolic curve optimization. Problems in these fields can be structured as optimization problems involving the maximization or minimization of some linear and non-linear parameterized objective functions. Therefore, finding problematic problems and searching for the best options have been priorities since the invention of technological devices. G. Dantzig, a USA air force employee, pioneered the numerical technique for solving linear programming problems in 1947 [1]. On this subject, Von Neumann developed the theory of duality [2]. Presently, thousands of algorithms have been developed for a wide range of applications, including optimization and problem solving. Optimization algorithms assist in finding an appropriate solution to a problem. There could be various approaches to a specific situation, but the best approach is the global one.

There are several examples of global optimization in engineering, financial services and management systems. In light of this and consequently, large-scale algorithms are separated into two categories: traditional algorithms, such as gradient-based algorithms or quadratic programming. Secondly, evolutionary algorithms include heuristic and meta-heuristic algorithms, as well as several artificial intelligence methods. Traditional algorithms are efficient and mostly deterministic throughout their operation. Moreover, they conduct local searches, so there is no guarantee of global optimality for the majority of optimization problems. As a result, traditional algorithms have limited solution diversity and cannot effectively solve multimodal problems because they do not work on highly nonlinear problems. Alternatively, Evolutionary Algorithm (EA) [3] for both heuristics and meta-heuristics could be a solution to previous limitations because they are stochastic. Additionally, meta-heuristic algorithms are more efficient than heuristic algorithms because heuristic algorithms rely on trial and fault to reach a solution.

Swarm intelligence[4] and evolutionary sophistication [5] both include meta-heuristic algorithms. EAs could mimic the ideas behind evolution in nature. The most effective and highly regarded algorithm in this class is the Genetic Algorithm (GA) [6], which is based on simulating Darwinian Theory of Evolution concepts [7]. Based on GA, Leo is an optimization problem that is formulated by starting with a particular problem's set of possible approaches. It modifies the variables of the solutions based on their fitness value after evaluating the solutions using the optimization problem. The random initial solutions tend to be highly influenced because the most competent individuals are more inclined to participate in improving other solutions. As a result, it is practically guaranteed that the initial random solutions will be improved. Other EAs are outlined in the literature such as, Evolutionary Programming [8], [9], Learner Performance-based Behavior (LPB) [10] and Quantum-based avian navigation optimizer algorithm [11] as well.

Many problems and applications that have been improved and reinforced in the field of nature-inspired [12] based meta-heuristics cannot be addressed in a reasonable time frame and at an acceptable computing cost. Because of the issue space's complexity or large number of variables, they use traditional methods or direct search techniques. Regardless of the circumstances, basic algorithm changes may be required to resolve the issue [13]. Effective stochastic optimization techniques are required for all such noisy objectives. The paper discusses stochastic single-objective optimization in high-dimensional parameter spaces. It is inappropriate for higher-order optimization techniques to be used in these cases, so we will only discuss first-order optimization methods. Since the discovery of these algorithms, plethora researchers have attempted to improve or use them to solve various problems in various fields [14]. The success of these algorithms in both business and science points to the value of bio-inspired-based techniques.

There are several factors contributing to the benefits of bio-inspired algorithms. The first point is that evolutionary optimization techniques remove the last information by generation, bio-inspired techniques store information about the search agent over the class of iteration. A second advantage of bio-inspired algorithms is that parameters are less restricted in bio-inspired algorithms and compared to EAs, bio-inspired algorithms have fewer operators. Lastly, the flexibility of bio-inspired techniques makes them easily adaptable to problems in various fields [15], [16], In this case, a collection of computational systems called Immunity Ratio was inspired by the biological immune system's defense mechanism. Besides, the immune tolerance ratio is the body's natural defense mechanism. It is composed of cells and molecules with the ability to recognize invading pathogens, bacteria, viruses, and other non-self-substances [17]. The results of this study show how to improve the albumin quotient in blood after vaccination using a bio-inspired based optimization process. It must concentrate on identifying a global point of immune function after receiving vaccination dosages to prevent the spreading of separate viruses, particularly after COVID-19. Additionally, when solving optimization problems, the Duality gap property of Lagrange Elementary is crucial to obtaining a deterministic solution. As a result, the main topic of the paper is the proposed innovative meta-heuristic algorithm based on GA. The dual convex Lagrangian functions [18] with discrete variables and the bio-inspired immunity system during vaccination are combined to propose a single-objective algorithm as the core objective of this sophisticated research.
The following are major aspects of the contribution to be proposed:

- During vaccination, the immune system makes use of an innovative bio-inspired intelligent algorithm that uses the albumin quotient of human blood. It employs a fitness function to generate appropriate weights because it provides quick convergence towards global optimality in terms of fair coverage of the search space. This aids the algorithm's exploration and exploitation phases. It also employs the albumin quotient and is a group-based algorithm. Leo is a general-purpose GA-based algorithm because it uses a similar operator to update individuals when both crossover and mutation operators are utilized to modify the structure to generate updated individuals.
- A novel optimum solution is developed using a novel algorithm operator based on the Lagrangian Dual Function (LDF) [19] that agrees with significant meta-heuristic optimization. It proves the Lagrange Elementary duality gap property, which is crucial for arriving at a deterministic solution.
- The selected population is sorted backwardly. The first sorted population has been divided into two subpopulations, and the group that was separated from the first group with the highest fitness is used. Priority is given to running the optimization process on the population groups with the best individuals first.

- In general, deterministic and stochastic optimizations find the first optimum value, but we need to determine the best value as well. Thus, Leo aim is to select the appropriate optimal point based on the LDF, which can find more than one optimal point and select the right one.
- Finally, Leo helps optimize performance and accuracy by addressing two novel real-application problems, focusing on pathological IgG in the nervous system and physical attacks in manufacturing systems.

The next remaining portions of the article are ordered as follows: Section 2 introduces related work that details the earlier EA. In additional, the next section highlights how the immune system from vaccinations, and then Lagrange Elementary affect the bio-inspired method in the follow section. Section 5 provides the mathematical models, the Leo algorithm, and its operators. In section 6, a thorough comparison of several benchmark functions is conducted to determine the appropriate algorithm focusing on experimental evaluations. The novel real-world case study is given to confirm and validate the accuracy of Leo algorithms in section 7. Ultimately, the work is concluded in Section 8, which also offers some suggestions for further research.

## 2. Background Theory

Meta-heuristic optimization algorithms started to take off in the 1960s [20]. In practical applications, meta-heuristic optimization techniques are commonly used to solve problems. Numerous algorithms take their main inspiration from biological, physical, and chemical phenomena. The majority of algorithms, like the gravitational search algorithm [21] and quantum algorithms [22] are dependent on the principles of gravity, the theory of general relativity, and interactions between things. Additionally, chemical algorithms emphasize the use of meta-heuristic algorithms that are based on reaction principles, such as chemical reaction optimization [23] and multiscale quantum algorithms for quantum chemistry [24]. This section provides additional information about our proposed method. The contents cover mathematical formulas, immunity system simulations based on vaccine distribution, and other associated procedures. The sections that follow explain how the Leo method is used to explore and exploit potential solutions to optimization problems.

It is confirmed that a significant part of optimization strategies involves swarm intelligence techniques that mimic the intelligence of natural swarms, groups, schools, or flocks of animals. In the domain of cellular robotics systems, Gerardo Benny and Joon Wang first presented swarm intelligence in 1989. After that, the area developed and the topic gained recognition [25]. Thus, ant colonies, bee colonies, bird flocking, eagle hunting, mammal herds, bacterial growth, fish schooling, and microbial intelligence are all developed examples of microbial intelligence. It is based on a biological specialization or group of collective behaviors of creatures that these algorithms are fundamentally based on. Without a centralized control system, some creatures can collectively ensure the survival of a colony. In other words, creatures obtain food on their own, even at lengthy distances from their nests or hives. This is without someone instructing them where to look for it or how to search effectively. The most widely exploited algorithms in this situation are Ant Colony Optimization (ACO) [26], Fitness Dependent Optimizer (FDO) [27], Dragonfly algorithm (DA) [28], Salp Swarm Algorithm (SSA) [29], Particle Swarm Optimization (PSO) [30], Whale Optimization Algorithm (WOA) [31], and FOX-inspired optimization algorithm [32]. To determine the shortest route to the nest or hive from a food source, the ACO and FDO algorithms imitate the individual interaction of ants and bees. The PSO algorithm models the group navigation and hunting behavior of birds. Other swarm intelligence methods described in the literature include: artificial bee colony [33], cat swarm optimization [34], cuckoo searches [35], bat algorithm [36], dolphin echolocation [37], elephant herding optimization [38], fruit fly optimization algorithm [39], grey wolf optimizer [40], and moth-flame optimization [41]. Therefore, swarm intelligence has a wide range of applications in anthropology, industry, technology, and basic research.

Additionally, several innovative Artificial Intelligence approaches based on the traits and behaviors of living species in nature have emerged in the past couple of decades, which are characterized as nature-inspired optimization algorithms or bio-inspired. Bio-inspired models are noted for their successful application to model optimization because they mimic natural biological behaviors. Among the most popular are biological algorithms based on swarm intelligence. These examples are Bio-inspired Optimization [42], COVID-19 Optimizer Algorithm [43], and Kidney-inspired Algorithm [44]. Likewise, algorithms that are based on nature or biological principles have found use in a variety of industries and professions. These include data mining, timetabling and scheduling, pattern recognition, manufacturing, engineering, economics, geoscience estimate rate and healthcare [45], [46], [47]. The stability or balance between exploration and exploitation has a significant impact on a meta-efficiency of meta-heuristics. The primary driving force behind the development of a novel optimization algorithm that is bio-inspired by the human immune system's power and authority of exploration and exploitation in the human and creature body. Several optimization strategies concentrate on this topic, but there is no hybridization between the method and truly difficult mathematics, such as the dual Lagrange function and the healthcare system.

## 3. Vaccination Immunity

In the enlightenment, not all algorithms derived from nature are biologically inspired, others are purely based on physics and chemistry. Many bio-inspired algorithms don't directly take advantage of swarming behavior. The term "bio-inspired" is preferred to "swarm intelligence-based" for this reason. For instance, genetic algorithms draw inspiration from nature rather than swarm intelligence. Differential Search Algorithm (DSA) [48] and Differential Evolution Algorithm (DEA) [49] are two algorithms that are challenging to classify. Since there is no obvious connection to any biological function, DEA cannot be considered biologically inspired in the true sense of the word. Additionally, DSA is an advanced evolutionary method for solving numerical nonlinear equations with real values. On the other hand, as it is parallel to GAs and uses the word "evolutionary", we categorize it as a bio-inspired algorithm for the time being. Biologically, it generates a large number of random combinations, that are subsequently assessed for fitness using immune system levels. The fittest pair produces new offspring, and the cycle continues until the global immune system generation set is produced.

Thus, the immune system protects the body from infection. When viruses or bacteria enter the body, they infect and proliferate. This invasion, also known as an infection, causes the disease. The immune system attacks this infection with white blood cells [50]. According to Figure (1) [51], macrophages, B-lymphocytes, and T-lymphocytes form the majority of these white blood cells. Intruders from outside the cells are attacked by B-cells, infected cells are attacked by T-cells, and macrophages allow the immune system to rid itself of invaders by absorbing foreign objects and activating an immune response [50].

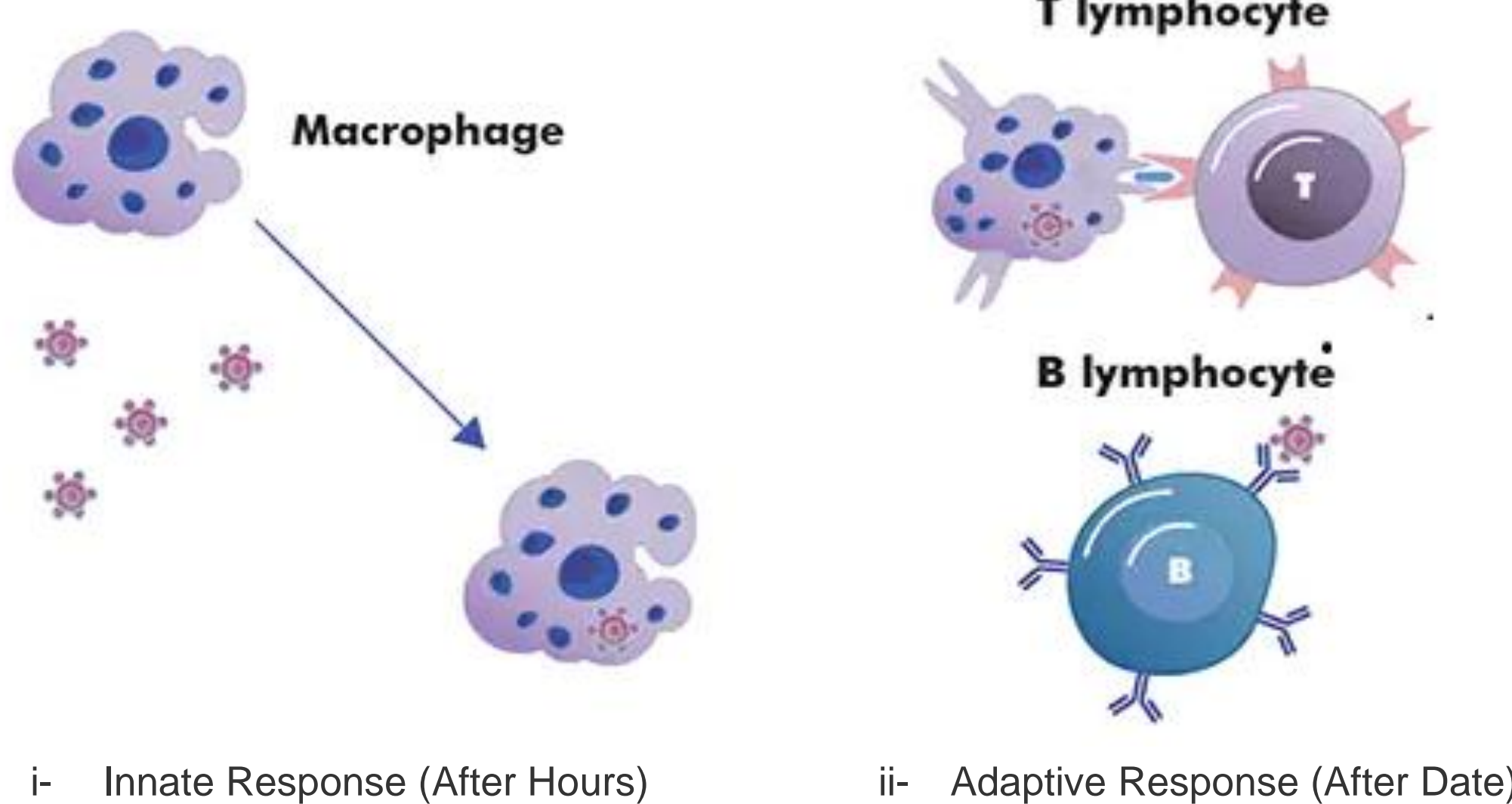

Figure 1. Response cells of the innate and adaptive immune systems. (Macrophages, B-lymphocytes and T-lymphocytes)

When calculating the defense system, it is important to consider the albumin quotient in human blood serum. To prove the immunity system is adequate, Immunoglobulin G ($IgG$) should be in the restricted range and read increasingly to demonstrate the immunity system is flawless. If the $\mathrm{Q}_{\mathrm{Alb}}$ (albumin quotient) is declining quickly and the $Alb_{\mathrm{serum}}$ is rising significantly, this indicates that the blood level of IgG is increasing and within the normal range [52]. If this is the case, the guardians have a high $Q_{Alb}$ and show that, according to function Eq.1 [53], humans have developed a modality-immunity system.

$$Q_{Alb} = \frac{Alb_{CSF}}{Alb_{serum}} \quad \text{Eq. 1}$$

Even when a person is still ill, behavioral traits and infectious diseases can spread through social interaction, particularly over the years that the COVID-19 epidemic has spread rapidly. It has been necessary to create effective vaccinations to maintain human immunity in order to stop this expansion. The immune system is naturally strengthened by vaccinations to combat sickness and reduce its effects. These technologies prevent the spread through a group and individuals copy their social connections for generating vaccination preferences. In this basic, this work examines the interaction between these two processes using bio-inspired algorithms and EAs based on complex mathematical functions. Accordingly, bio-inspired algorithms can be called a population-based algorithms. Liang and Cuevas-Juarez first suggested the virus optimization algorithm in 2016, and Liang et al. later enhanced it [54]. The outcomes of its application, like those of many other meta-heuristics, highly rely on its starting configuration. Without providing customized attributes for specific viruses, it simulates basic viruses. Its value goes beyond any argument, as demonstrated by the results. Nevertheless, Leo relies on generating new individuals by LDF to locate global points to mimic generic viruses for this purpose.

These proposed previous models cannot accurately account for the clustering of vaccination practices in a group of individuals since it requires that the group is evenly mixed. As a result, the concentration of anti-vaccination attitudes can increase disease outbreaks by interfering with protective immunity [55],[56]. To assess how imitation dynamics affect vaccination rates and disease outbreaks. The algorithms make models that determine the optimal global decision, duplicating individual behavior to create the highest-quality immune system for human vaccination. To confer protection; however, various vaccinations function in multiple ways. For instance, COVID-19 vaccines support disease prevention by assisting our bodies in developing immunity to the COVID-19 virus. Different vaccine types protect in different ways, such as mRNAs, viral vectors, protein subunits, and inactivated vaccines [57]. In general, all vaccines are eliminated in the body with a reserve of "memory" T-lymphocytes and B-lymphocytes capable of countering the virus in the future. As a result, these vaccinations promote immune response by raising, generating antibodies, and creating memory cells that will recognize and react if the body contracts the actual virus. Figure (2) illustrates the steps in developing an immune system toward spike proteins [58].

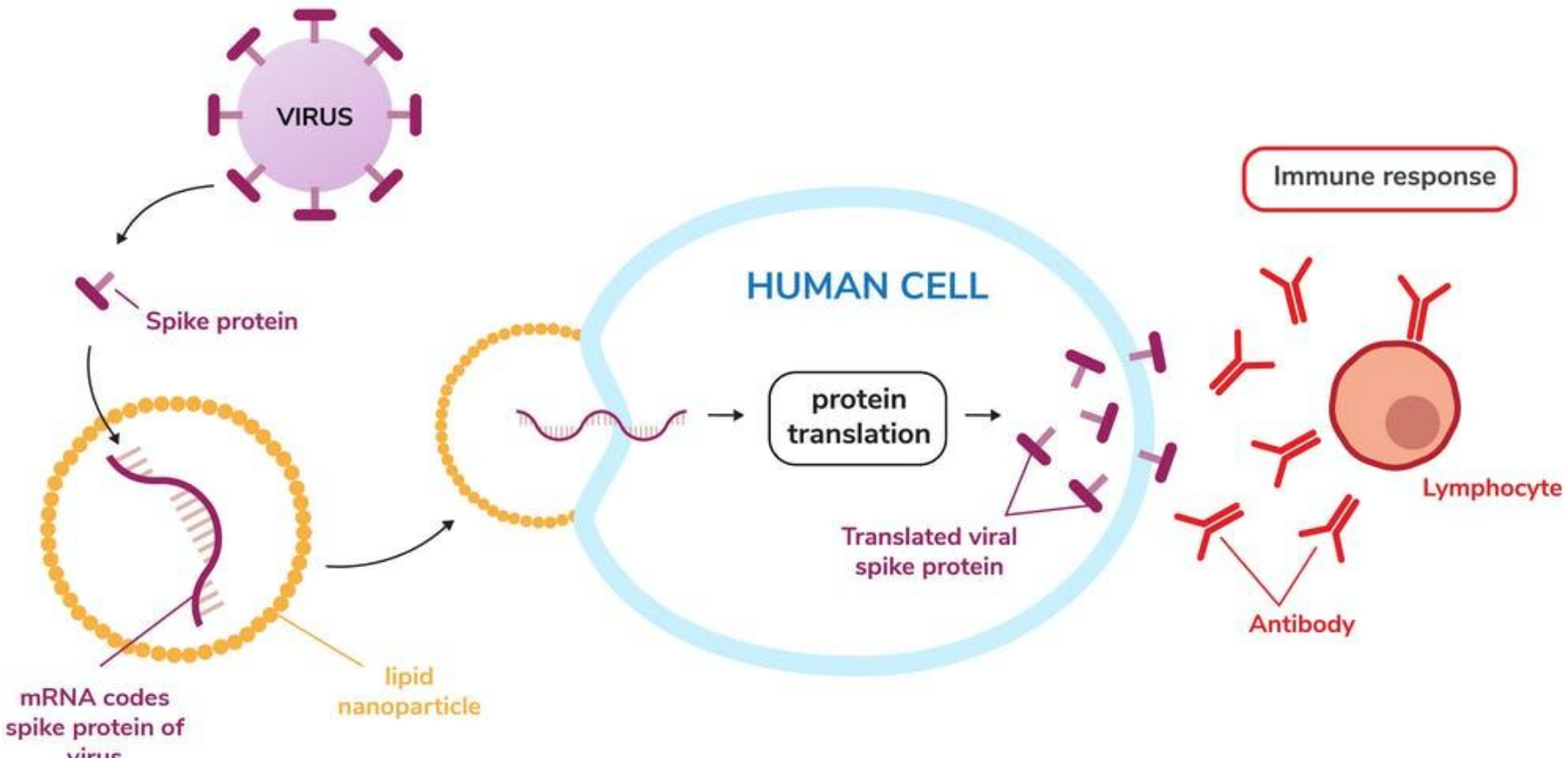

Figure 2. example of spike protein (mRNA) vaccine cycle life

## 4. Lagrange Elementary for Optimization

An evolutionary Lagrange approach to handling double-precision or single-precision restricted optimization problems. The Langrage theorem can be used in a variety of ways to find several optimal solutions. The mixed-numbers constrained optimization issue is converted into an unconstrained minimum or maximum problem with decision variable minimization and Lagrange multiplier maximization using a variety of methods. To implement the evolutionary Lagrange technique, a method like Lin et al. (2003) integrated into the evolutionary min–max algorithm [59]. This EA combines self-adaptation for penalty parameters to achieve global integration, allowing for the use of smaller penalty parameters without impacting the final search results. Besides, the Lagrange interpolating polynomial is the solitary polynomial of the lowest degree that interpolates a given collection of data in numerical analysis and data mining. For mathematicians, the foundations of mathematics are a dynamic complex system. A method for solving discrete optimization problems relied on the interpretation of calculus algorithms. As a result of the branch dynamics of the complex system and the stochastic algorithm that is still undetermined, it is challenging to manage the collective dynamics [60].

EA seems to be a useful tool for optimizing complex and non-convex optimization problems. The earlier algorithms, such as Suma (2021), employing EA-based network reconstruction methods in the proposed new algorithm [61], have concentrated on the broad applicability of EA-based methodologies and their enhanced efficiency in solving reconstruction situations in complicated applications. Also, both the objective and constraint functions are non-convex and incorporate expectations over random states to find a stationary point for a broad non-convex stochastic optimization problem. In addition, Liu et al. (2019) presented a constrained stochastic successive convex approximation algorithm [62]. In several deliberations, it was demonstrated that using a hybrid algorithm and Lagrange technique to solve limited optimization problems. Using a single generation to develop the most optimal solution, the mixed-integer hybrid differential evolution algorithm [63] was established. It developed a straightforward way to encode mixed-integer variables. An effective and reliable method is developed to solve the mixed-integer restricted optimization problem as developed in the Leo algorithm for more than dual generations of the enhanced Lagrange function. Thus, it is essential to understand that Leo is a type of genetic optimization that modifies parent genes to update the search agent position. To successfully navigate the search space, the population is divided across many albumin quotients through the suggested Lagrange orientation and Lagrange multiplier stationary point navigation based on crossover strategies and mutation strategies.

In general, the Lagrange multiplier technique must be used to maximize or minimize a multivariate function; the input includes any number of dimensions of $f(x, y, \lambda)$,, and which typically takes the form of another multivariate function $g(x, y)$ becoming set identical to a constant ($c$) [19]. Using gradients ($g$) as a two-variable function, this issue shows how to achieve the best solution for reaching the top of the cliff side in Figure (3).

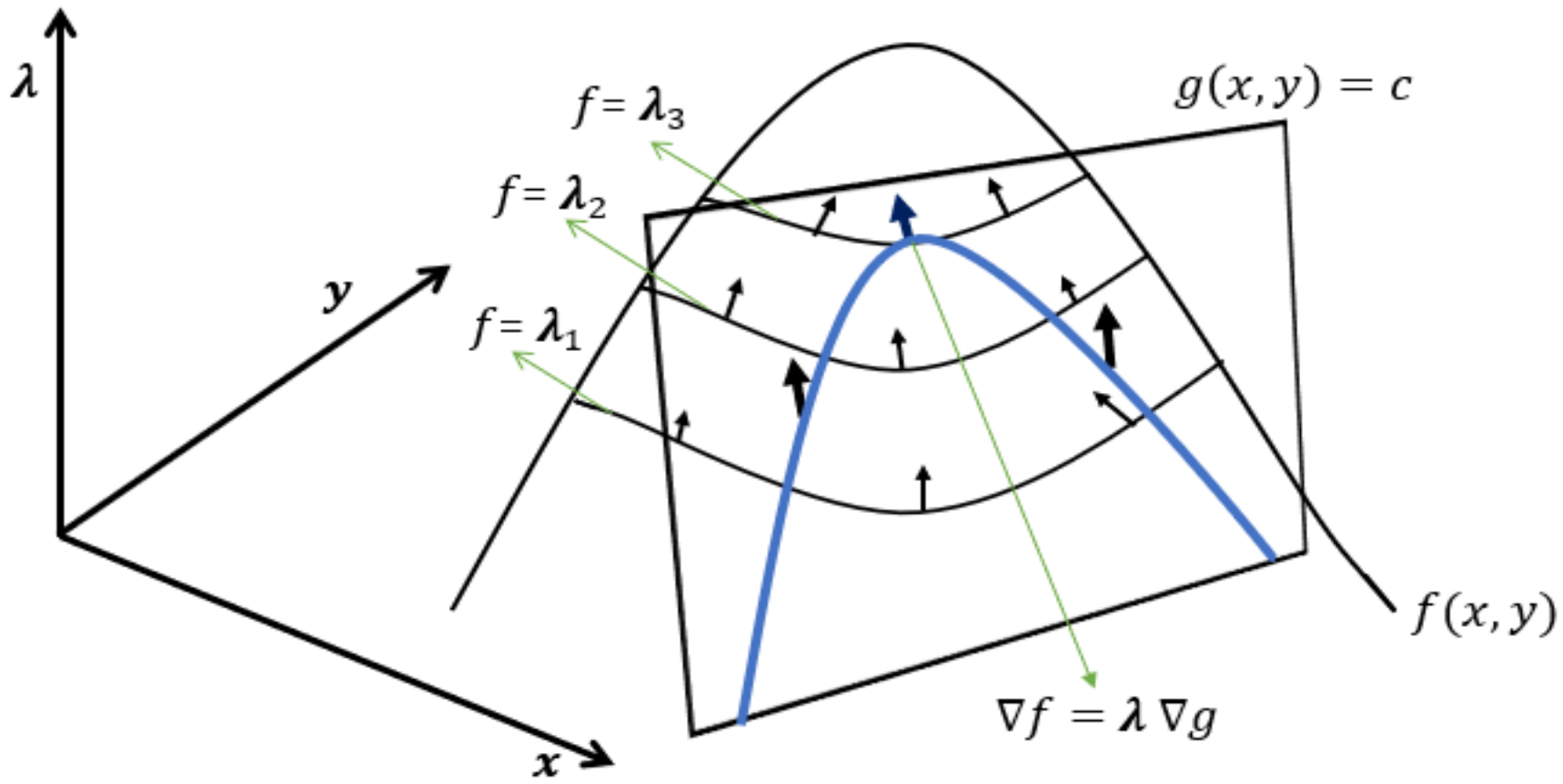

Figure 3. Given that the solution can't ascend significantly higher than the point where the restriction $g = c$ crosses the top, the objective is to climb as high on the top as possible using the Lagrange theorem.

As Eq. 2 with a derivative for multipliers, $F$ is known as the Lagrange function and the numbers $\lambda_0, \lambda_1, \ldots, \lambda_m$ as Lagrange multipliers. If $g_i(y)$ is a regular value of the map $g = (g_1, \ldots, g_m)$, the statement takes on a more elegant form. This method is particularly helpful for the first and second theorems of Lagrange multipliers because it frequently allows for the solution of the associated conditions without the need for explicit formulas that express point set accumulation in terms of $(n-m)$ independent variables. In general, a private blockchain of relations, or a system of $(n+m)$ equations in $(n+m)$ variables, is frequently formed by the requisite conditions established by using a Lagrange function.

$$F(\lambda, x, y) = f(x) + \sum_{i=0}^{m} \lambda i\big(gi(y) - gi(x)\big) \qquad \text{Eq. 2}$$

$$When \quad \frac{\partial F}{\partial x_j}(x^*, \lambda^*) = 0 \qquad \forall_j \in \{1, \ldots, n\}$$

$$And \quad \frac{\partial F}{\partial x_i}(x^*, \lambda^*) = 0 \qquad \forall_i \in \{1, \ldots, m\}$$

It may often summarize these criteria; it aims to look for constants $x_0, y_0$ $and$ $\lambda_0$ that fulfill $g(x_0, y_0) = c$. Depending on the requirements, Eq.3 illustrates the tangency conditions.

$$f(x_0, y_0) = \lambda_0 \nabla g(x_0, y_0) \qquad \text{Eq. 3}$$

This can be broken into its components as Eq.4 and Eq.5:

$$f_x(x_0, y_0) = \lambda_0 \nabla g_x(x_0, y_0) \qquad \text{Eq. 4}$$

$$f_y(x_0, y_0) = \lambda_0 \nabla g_y(x_0, y_0) \qquad \text{Eq. 5}$$

Lagrange Eq. 6 is developed as a completely distinct function from the broken tangency conditions that accepts all the same inputs as $f$ and $g$ as well as the new kid in the action, which is now treated as a variable rather than a constant when $c = 0$. The function, associated with the Lagrange multiplier concept, is used to develop prerequisites for conditional maxima or minima of functions with many variables or, in a more general context, of functionals. The main purpose is to locate local maxima (respect minima) or local minima (respect maxima).

$$l(x, y, \lambda) = f(x, y) - \lambda(g(x, y) - c) \qquad \text{Eq. 6}$$

According to GA, individuals must generate different individuals by modifying suggested self-adaptive systems during each phase of development. A crossover parameter, for example, can boost the global search capability and increase the variance in the differential vector. A rounding procedure, on the other hand, reduces the second component of the weighted difference vector to the closest individual number [64]. Thus, the updated vector now consists of fresh individuals. Therefore, it is simple to implement the evolution of parameters. The resulting crossover vector may occasionally exceed the search restrictions. In this case, reducing the value of the boundaries is necessary. In GA optimizations, this action is also permitted for the mutation operator. The parameters are improved using EA based on Lagrange models. To determine the relationship between the individual vertices, the structure of the gene from the second known individual is employed [65]. Additionally, complex parameter problems like heuristic games [15], biology systems or physical problems [66], fault stability framework [67] and educational case statements[68] can be further simplified to retrieve the genetic framework. Simply stated, these genetic operators are used to create new equations from Lagrange Tangency Conditions. The subsequent sections of the Leo algorithm provide the perfect illustration of such a problem.

## 5. Leo Algorithm Deterministic Process

The primary search operators in the bio-inspired vaccination immunity algorithm are crossover and mutation, which are key components of evolutionary algorithms (EAs). The crossover operator creates new offspring by merging the genetic information of two parent solutions. This process is guided by the principles of the Lagrange multiplier problem, which helps optimize the search for the best solutions. This mimics natural reproduction, where the fittest individuals, determined by their immunity levels, are chosen to create the next generation. The mutation operator adds random variations to the offspring, akin to genetic mutations, which helps the algorithm explore new areas of the search space. These mutations alter the positions of the search agents, allowing the population to traverse the search space more efficiently. The crossover and mutation approaches are critical for optimizing the global decision-making process, ultimately leading to an enhanced immune system model for vaccination. The combination of Lagrange multiplier techniques further refines these operators to minimize the objective function under given constraint.

Thus, the specific function minimized by the Leo algorithm is clearly identified in Equation 1, aiming to optimize the immune system during vaccination by minimizing the $\mathrm{Q_{Alb}}$ in the context of human or animal samples. The algorithm primarily focuses on this parameter to determine the global solution within blood serum. This approach illuminates the process by which a genome is generated and iteratively refined until a new optimal solution is identified, ultimately achieving global optimization points.

As a result, this algorithm mimics the actions of a swarm of immune systems from a group of people during imitation. The core of this method was inspired by parents' efforts to find an appropriate, compatible group of people among numerous candidate groups, which contains $Alb_{\mathrm{serum}}$ in human blood. In addition, choosing the most effective immune system $(IgG)$ among multiple positive systems is thought to be convergent to optimality. Every genome that looks for new groups of parents have a high $\mathrm{Q_{Alb}}$ provides a latent solution to this algorithm. The algorithm starts by initializing an $Alb_{\mathrm{serum}}$ the population at random in the search space or population $\mathrm{X}_{ţ,i}(i = 1, 2, \ldots N)$; where ţ is a step to initialize selection and each genomic position indicates a newly discovered $\mathrm{Q_{Alb}}$ (solution). Since this algorithm seeks out the highest-quality parents within a stochastic space through a group-based population approach, the first step is to select individuals from the population. For further clarity, Figure 4 presents a flowchart that illustrates the step-by-step process of the Leo algorithm. This flowchart visually explains the various stages involved in optimizing solutions and achieving convergence through two distinct approaches: a simple process schema shown in Figure (4-B), and a complex schema depicted in Figure (4-A).

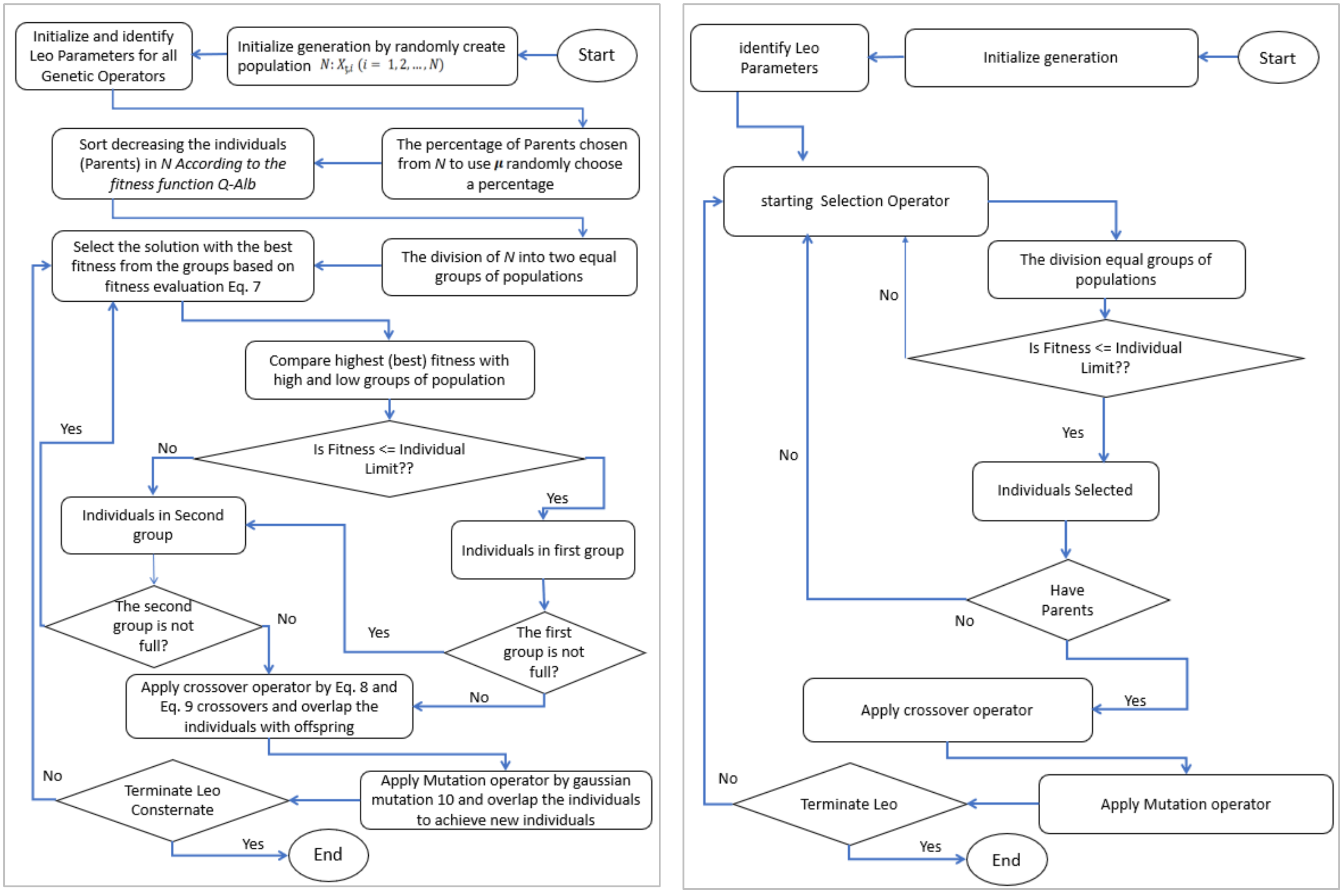

A- Leo Complex Process Schima B- Leo Simple Process Schima

Figure 4. Leo algorithm flowchart process

In the first part of the textual Leo algorithm pseudocode shown in Figure (5), all Leo parameter settings and symbols have been identified. For further identification, Table (1) outlines all the parameter settings for the Leo algorithm and covers the specific qualifiers for each parameter. Additionally, if the termination condition is not met, a percentage of individuals will be selected at random from $N$ as a parameter. Additionally, all the symbols required to process this algorithm are identified in Table (2) for clarification of the Leo process represented in Figure (5).

Table 1. Leo Algorithm Parameter Settings.

| **Parameters** | **Parameter Value** |
|---|---|
| Population Size | 100 |
| **μ** the percentage of individuals chosen from $N$ | 0.5 |
| Number of Offspring | 2*round (Crossover rate * Population Size /2) |
| Crossover rate | 0.6*population size |
| Mutation rate | 0.3*population size |

Initially, the population size is limited to 100 individuals as mentioned in Table (1). GA states that the best individuals should be selected even before passing them on to new operators to generate new best genomes. This stage is crucial since it requires splitting the main population into two equal subpopulations after appropriately sorting the population in descending order. Then, the primary Half Group (ɦɠ) also divided into two random groups: First Group ($f$ɠ) and Second Group (ʂɠ) pointed out in Table (2). At that point, the individuals will be selected from the half-group-populations based on the fitness function Eq.7 to find the most optimal $Q_{Alb}$ from $Alb_{serum}$. It should be select the highest fitness from individuals of (ɦɠ) identified by ($X^{*}_{ţ,i}$) then it will compare all individuals from $\text{ɦɠ}_{\text{ʂɠ highest fitness}}$ and $\text{ɦɠ}_{f\text{ɠ: highest fitness}}$ from $\text{ɦɠ}_{\text{highest fitness}}$.

$$X_{ţ,i+1} = \frac{X_{ţ,i}}{X^{*}_{ţ,i} \quad for\ \text{ɦɠ}_{\text{highest fitness}}} \qquad \text{Eq. 7}$$

Because selected individuals can originate from the first portion of the sub-population, this eliminates convergence to local optima [69]. The final two phases are used to enhance the individuals by allowing the genes to operate in groups. This requires support from other genes as well as the individuals used as parent selection which is prepared to create the proper next steps. The best weight of metacognition generally influences how a genome investigates issues (mutations). However, when individuals work together, their respective genomes have an impact on how they behave during the process (crossover). After the acceptance of the genomes in the $Alb_{serum}$ from human blood to estimate the $Q_{Alb}$, as indicated, they may have an effective ratio of the human immune system. As a result, they may benefit from boosting the $IgG$ rate through vaccinations such as offering assistance and collaborating with others. Furthermore, as mentioned in [70], genomes might affect one another's frequency. They can interact with each other on gene groups in blood serum or request assistance when vaccine doses impact the genome. As previously stated, the Leo algorithm is based on GA. The GA operators mimic the process of gene heredity to create original individuals at each generation. As shown in the illustration, the operators are used to modify the structure of individuals. Selection, crossover, and mutation are the most common genetic operators. In Leo, genes are used as selection operators, while Leo focuses on group-based selection. Following that, we explain how the Leo algorithm works while crossover and mutation operators are utilized.

Table 2. Leo symbol identification in current process

| **Identifier** | **Main Definition** |
|---|---|
| $N \leftarrow 0$ | The original 'initial' random population |
| $X \leftarrow 0$ | The number of Parents in the novel population |
| $\mu \leftarrow 0.5$ | The percentage of Parents chosen from $N$ |
| $ţ \leftarrow 0$ | Iteration Step Initialize selection in Leo starting |
| ɦɠ | half-group-population chosen from $N$ in the end of iteration |
| $f$ɠ | first-group- population |
| ʂɠ | second- group- population |
| $k$ | calculate the number of newly created parents (Individuals)by counter |

```
1.  Algorithm: Leo (N, X, μ, ţ, ɦɠ, fɠ, ʂɠ, k)
2.  Table 2 provides Leo symbols identification clarification.
-----------------------------------------------------------------
3.  //1- Initialize generation
4.  Randomly create a population N: X_{ţ,i} (i = 1, 2, … , N)
5.  //2- Identify parameters, crossover rate and mutation rate
6.  Identify genomes X for all Q_Alb [albumin quotient]
7.  to get high IgG [best immunity system]
8.  //3-Selection Groups of genomes
9.  do {
      a. Use μ parameter to randomly choose a percentage
      b. Evaluate the fitness of individuals in X by Eq.7 by fitness function
      c. sort decreasing the individuals (Parents) in N
      d. Select ɦɠ: N/2 [Divide N to two equal parts of populations]
      e. // generate two groups from ɦɠ: fɠ and ʂɠ
10. determine the best fitness evaluation in first group by Eq.7
      a. //compare the highest fitness with high and low populations
      b. while k <= X
           i. if fɠ include parents
                1. if X_ţi from ɦɠ_highest fitness <= ɦɠ_fɠ: highest fitness
                2. collect X_ţi in the fɠ
                3. else
                4. collect X_ţi in the ʂɠ
                5. end if
           ii. else
           iii. collect X_ţi in the ʂɠ
           iv. end if
           v. k ← k+1;
      c. end while
11. //4- Lagrangian Problem Crossover LPX
12. Required: Implement Eq.8 and Eq.9
13. swapping first genome to find new individuals according Figure 6
14. //5- Mutation [Gaussian Mutation]
15. //Look at 5.2 section
16. ţ← ţ+1;
17. } while fitness of fittest individual in ɦɠ is not high enough
18. [Select the best solution from the ɦɠ];
```

Figure 5. Leo Algorithm Pseudocode

In the final discussion, the Lagrange multiplier proves to be a powerful tool in Leo algorithms, particularly by incorporating constraints into the objective function during selection and crossover operations. By introducing Lagrange multipliers, it converts a constrained problem into an unconstrained one, simplifying the problem-solving process and helping discover new optimal solutions, as demonstrated in Eq. of 6, 7, 8, and 9, and shown in Figure 3. This method is vital for efficiently minimizing the objective function while ensuring all constraints are met, as detailed in the subsequent subsection.

### *5.1 Leo Crossover Process*

The crossover operator from GA-based is used in the Leo algorithm to demonstrate this stage. Utilizing a crossover operator will facilitate the exchange of some genes between genomes for achieving a perfect immune system after vaccination. In consequence, the genome is an entire set of genes that are different from the original genomes of the $Alb_{\text{serum}}$. As a result, the overall genome of both individuals will be affected, and the produced individuals will have a different $Alb_{\text{serum}}$. Generally, the study clarifies that crossover concepts within the Leo algorithm can be either essential components or optional strategies, depending on their request. It details how these crossover operations contribute to the algorithm's agenda, either by integrating directly into the core process or by functioning independently. The role of crossover is emphasized as crucial for generating new genomes, which allows the algorithm to explore different areas of the solution space. This exploration is dynamic for identifying new optimal values in each iteration, as crossover helps refine the search process based on the fitness function.

As Adam optimization pointed out [71]; Stochastic gradient-based optimization is crucial in several fields of science and engineering. Multiple gradient theorems can be used to determine whether a function is differentiable by Lagrange. Gradient descent is a fairly effective optimization method when its parameters are lagged. Thus, Leo is applied to Lagrangian Problem Crossover (LPX) [19] As discussed in Sections 1 and 2, the standard approach involves generating inclusive individual structures by seeking alternative genes. This method finds new solutions efficiently by producing two new offspring, which provide a direct solution to the problem at the given rate. The Lagrange Multiplier plays a significant role in this process. It helps find optimal solutions, as pointed out in Eq. 8 and 9, in constrained optimization problems by converting them into a system of equations that balance constraints and objectives. This algorithm begins with a rate of crossover equal to 0.6 multiplied by the population size divided by two. As noted in Section 2. The LDF theorem is based on real-world equation examples in the LPX standard. It would be proposed a quite novel crossover operator that would identify several local points. The LDF theory is an advancement that serves as an alternative to the Conic Duality hypothesis, building upon the Lagrangian function concept, as discussed in the previous sections. Eq. 8 and Eq. 9 [19] are used to find new structural individuals ($O_{ţ,i}$).

$$O_{ţ1} = \left(x_{ţ1} - x_{ţ2}\right)^2 + \left(x_{ţ2} - 1\right)^2 - \Big(\alpha\left(x_{ţ1} + 2x_{ţ2} - 1\right) + \alpha\left(2x_{ţ1} + x_{ţ2} - 1\right)\Big) \qquad \text{Eq. 8}$$

$$O_{ţ2} = \left(x_{ţ2} - x_{ţ1}\right)^2 + \left(x_{ţ1} - 1\right)^2 - \Big(\alpha\left(x_{ţ2} + 2x_{ţ1} - 1\right) + \alpha\left(2x_{ţ2} + x_{ţ1} - 1\right)\Big) \qquad \text{Eq. 9}$$

A Lagrange Multiplier is a random value for LPX that has been generated by the Lagrange Multiplier. As a result, it proves the existence of deterministic properties for the Lagrangian dual function and the Leo problem optimization. The values come from a uniform distribution on the interval [0.2, 0.3] in this algorithm. It is possible to exchange the first old gene for the second new gene when new individuals are created. This will determine the updated fitness function, and then determine the fitness using Eq.1 additional declaration and in Figure (6), the Leo crossover has been constructed using pseudocode.

$X_{ţ,k}$, $X_{ţ,K+1}$: are the two given Genes;
$O_{ţ,k}$, $O_{ţ,k+1}$: are the two new Offspring;
$Y_{ţ,k}$, $Y_{ţ,k+1}$: are the two highest fitness Genes
$\alpha$ : is a random value between (0.2,0.3);
$k$: is counter;
i: is additive by one gene number
$n$: number of parents (individuals)

**Leo_Crossover ($X_{ţ,k}$, $X_{ţ,K+i}$, $Y_{ţ,k}$, $Y_{ţ,k+i}$){**
**Required:** Calculate $\alpha$ value
**While** $k$ smaller than $n$
    Calculate the first offspring $O_{ţ,k}$ from eq.8;
    Calculate the second offspring $O_{ţ,k+i}$ from eq.9;
    Calculate the new first gene: $Y_{ţ,k} = \frac{O_{ţ,K+i}}{X_{ţ,k}}$;
    Calculate the new second gene: $Y_{ţ,k+i} = \frac{O_{ţ,k}}{X_{ţ,K+i}}$;
    $k + i + 1$;
**End while**
} the best individuals found during the evaluation

Figure 6. Leo Crossover Process Pseudocode

### *5.2 Leo Mutation Process*

The number of immunological ambiguities caused by a genetic mutation may be explained, however, it is still unidentified to scientists. Another random effect of mutation occurs when it triggers behavioral changes in various individuals. The most basic form of mutation involves modifying one or more genes. Metacognition can also have a stochastic influence on the overall behavior of genes, as formerly mentioned. By adjusting the levels of immunity-related activities in their genes according to their mutation rate, individuals can alter their behavior in a specific route. The mutation operator from GA-based algorithms is often used to illustrate this development within algorithms. In the context of the Leo algorithm, the role of mutation is crucial for discovering solutions and improving the search process in each iteration. While not all algorithms depend on mutation, particularly those rooted in GA, mutation introduces diversity by randomly altering gene values. This randomness can simplify the discovery of new solutions during iterations, although it may infrequently lead the search away from the optimal solution. However, in most cases, especially when testing the Leo algorithm, mutation proves effective in efficiently uncovering new solutions.

The main goal of EA mutation is to start introducing diversity into the total population sample. Mutation operators are used to avoid local minima or local maxima by preventing the population of genomes from becoming excessively similar to each other. This results in slowing or even halting convergence to obtain the global optimum. In stochastic algorithms, every point in the search space should be reachable by one or more mutations, which means that small mutations might have more of a significant impact than large ones [72]. Individuals can be used in a different situation or to serve genes that are not present in the initial population. Individual representations can be mutated in a several different ways. For instance, uniform mutation, replacement mutation, scramble mutation, inversion mutation, dynamic mutation, boundary mutation, and others. Leo utilizes Gaussian Mutation [73] as the operator for EA. Self-adaptation is the ability of a GA's ability to modify its algorithm while solving a specific problem [74]. Since it has been demonstrated that the Gaussian mutation operator is the most effective and popular option for self-adaptation in GA. Thus, when sigma ($\sigma$) is a random value between (-1,1) and $(j_i)$is a random sample, Eq. 10 is used to develop a Leo mutation operator $M_{t,i}$ . Whenever the Leo percentage mutation for individuals' sample $X_{t,i}\,(j_i\,)$ equals to (0.3).

$$M_{t,i} = X_{t,i}\,(j_i\,) + \sigma * randn\big(size\,(j_i)\big) \qquad \text{Eq. 10}$$

## 6. Result and Discussion

To make sure the suggested algorithm works correctly and determine the effectiveness of this algorithm, several common benchmark functions from the literature are used. A further five well-known algorithms from the literature are compared with Leo algorithms results 19 classical benchmark tests; one of them is more popular, such as DA, PSO, and GA; two others are novel. Besides, the results of the proposed algorithm are compared to Leo for CEC-2019 test functions such as DA, WOA, SSA, FDO, LPB, and FOX. Then, these results are statistically applied to each other to prove whether the result is significant using the Wilcoxon rank-sum test. Therefore, to ensure that the Leo works well to solve real-world applications, most of the parameters are defined in the Leo pseudocode in the previous section. Additionally, the results of CEC-C06 2019 test functions such as DA, WOA, SSA, FDO, LPB, and FOX are compared to Leo. The Wilcoxon rank-sum test is then used to statistically compare these results to each other and determine whether the result is significant. The majority of parameters are defined in the Leo pseudocode in the previous part. Finally, the Leo functions have been successfully used to solve real-world applications.

### *6.1. Classical Benchmark Test Functions*

The Leo algorithm has just been tested 30 times with 100 search agents; within every test, the algorithm looked for the most efficient optimum solution in 500 iterations, and then the average (AVA) and standard deviation (STD) were calculated. As discussed in Table (3 and 4), the DA, PSO, GA, FDO and LPB parameter sets are described in this paper. For this general convex learning problem, our results are comparable to the best-known bound. To evaluate the effectiveness of the Leo algorithm, three sets of test functions are selected. Unimodal, multimodal, and composite test functions are the three different categories into which the test functions have been categorized [75]. Each of these test functions is designed to evaluate the algorithm's effectiveness and benchmark specific perspectives. For instance, unimodal benchmark functions are used to verify the algorithm's exploitation level and convergence because their name indicates that they have a single optimal. Multimodal benchmark functions contain multiple optimal solutions to test local optimum avoidance and exploration levels. There are many optimum solutions, just like in multimodal algorithms; among them are the global optimum solution and the majority of individual optimum solutions. To get globally optimal solution, an algorithm must avoid local optimum solutions. In addition, the majority of composite benchmark functions are blended, rotated, shifted, and biased versions of other test functions. Composite benchmark functions contain a very large number of local optima and offer a variety of shapes for various search landscape regions. This kind of benchmark function is illustrated by Tables (9, 10, and 11) [76] in the appendix.

Leo algorithm was run in accordance with the fundamental steps previously indicated, and the resulting findings were compared to those of DA, PSO, and GA, three well-known alternate algorithms whose results are reported in papers [77] found in Table (3). Thus, DA produced optimal results in the first unimodal function and PSO performed more efficiently in the sixth test function. According to the results of tests TF2, TF3, TF4, TF5, and TF7, Leo generally delivered better performance than the other algorithms for exploitation capacity and optimum results. Additionally, the Leo algorithm produced the same superior results when compared to the FDO and LPB algorithms whose results are reported in papers [10], [27], except for TF1 and TF6 showing better exploitation level and convergence which are stated in Table (3). Leo also outperformed the other algorithms in TF11 and TF13 and achieved the second rank, the only one with a superior result in TF12. When compared with all algorithms listed in Table (3) except TF11 and TF13, Leo achieved the second rank for these two functions. Concerning the multimodal functions, Leo attained the second rank with a superior result in only TF12 compared with the other algorithms in TF11 and TF13. Leo, on the other hand, outperforms all other algorithms in Table (3) with the exception of TF9 and TF10, where they have earned the second position. In composite test functions, for every test function, Leo performed more efficiently than DA, PSO, and GA. In contrast, Leo algorithms consistently underperformed LPB algorithms. In all test functions, Leo achieved the second rank with better performance than FDO; see also Tables (3 and 4).

Table 3. Comparing the results of Leo with DA, PSO and GA algorithms on classical test functions

| TF | **Leo** | | **DA** | | **PSO** | | **GA** | |
|---|---|---|---|---|---|---|---|---|
| | AVA | STD | AVA | STD | AVA | STD | AVA | STD |
| TF1 | 2.69874E-09 | 7.49992E-09 | **2.85E-18** | **7.16E-18** | 4.20E-18 | 1.31E-18 | 748.5972 | 324.9262 |
| TF2 | **3.7305E-06** | **3.95635E-06** | 1.49E-05 | 3.76E-05 | 0.003154 | 0.009811 | 5.971358 | 1.533102 |
| TF3 | **5.31468E-09** | **2.07901E-08** | 1.29E-06 | 2.10E-06 | 0.001891 | 0.003311 | 1949.003 | 994.2733 |
| TF4 | **3.60286E-05** | **3.22842E-05** | 0.000988 | 0.002776 | 0.001748 | 0.002515 | 21.16304 | 2.605406 |
| TF5 | **10.60296667** | **13.93285916** | 7.600558 | 6.786473 | 63.45331 | 80.12726 | 133307.1 | 85007.62 |
| TF6 | 4.31581E-10 | 5.51803E-10 | 4.17E-16 | 1.32E-15 | **4.36E-17** | **1.38E-16** | 563.8889 | 229.6997 |
| TF7 | **0.001449721** | **0.002690575** | 0.010293 | 0.010293 | 0.005973 | 0.003583 | 0.166872 | 0.072571 |
| TF8 | -2989.147333 | 202.684514 | -2857.58 | 383.6466 | -7.10E+11 | 1.2E+12 | **-3407.25** | **164.478** |
| TF9 | 37.07867 | 12.2775166 | 16.01883 | 9.479113 | **10.44724** | **7.879807** | 25.51886 | 6.66936 |
| TF10 | 4.8836E-05 | 2.89869E-05 | **0.23103** | **0.487053** | 0.280137 | 0.601817 | 9.498785 | 1.271393 |
| TF11 | **2.7393E-08** | **5.51514E-08** | 0.193354 | 0.073495 | 0.083463 | 0.035067 | 7.719959 | 3.62607 |
| TF12 | 1.87667E-08 | 2.89749E-08 | 0.031101 | 0.098349 | **8.57E-11** | **2.71E-10** | 1858.502 | 5820.215 |
| TF13 | **8.90491E-09** | **1.88063E-08** | 0.002197 | 0.004633 | 0.002197 | 0.004633 | 68047.23 | 87736.76 |
| TF14 | **6.9979** | **5.833242622** | 103.742 | 91.24364 | 150 | 135.4006 | 130.0991 | 21.32037 |
| TF15 | **0.001673093** | **0.003539145** | 193.0171 | 80.6332 | 188.1951 | 157.2834 | 116.0554 | 19.19351 |
| TF16 | **-0.622100333** | **0.396782974** | 458.2962 | 165.3724 | 263.0948 | 187.1352 | 383.9184 | 36.60532 |
| TF17 | **1.788405333** | **2.237631581** | 596.6629 | 171.0631 | 466.5429 | 180.9493 | 503.0485 | 35.79406 |
| TF18 | **3.590623333** | **0.711917144** | 229.9515 | 184.6095 | 136.1759 | 160.0187 | 118.438 | 51.00183 |
| TF19 | **-2.670808** | **1.185307969** | 679.588 | 199.4014 | 741.6341 | 206.7296 | 544.1018 | 13.30161 |

Table 4. Comparing the results of Leo with FDO and LPB algorithms on classical test functions

| TF | Leo | | FDO | | LPB | |
|---|---|---|---|---|---|---|
| | **AVA** | **STD** | **AVA** | **STD** | **AVA** | **STD** |
| TF1 | 2.69874E-09 | 7.49992E-09 | **7.47E-21** | **7.26E-19** | 0.001877545 | 0.002093616 |
| TF2 | **3.7305E-06** | **3.95635E-06** | 9.39E-06 | 6.91E-06 | 0.005238111 | 0.003652512 |
| TF3 | **5.31468E-09** | **2.07901E-08** | 8.55E-07 | 4.40E-06 | 36.4748883 | 29.22415523 |
| TF4 | **3.60286E-05** | **3.22842E-05** | 6.69E-04 | 0.0024887 | 0.393866 | 0.135818 |
| TF5 | **10.60296667** | **13.93285916** | 23.501 | 59.7883701 | 16.76919 | 22.19251 |
| TF6 | 4.31581E-10 | 5.51803E-10 | **1.42E-18** | **4.75E-18** | 0.00203173 | 0.0027832 |
| TF7 | **0.001449721** | **0.002690575** | 0.544401 | 0.3151575 | 0.004975 | 0.002965 |
| TF8 | **-2989.147333** | **202.684514** | -2285207 | 206684.91 | -3747.65 | 189.0206 |
| TF9 | 37.07867 | 12.2775166 | 14.56544 | 5.202232 | **0.001567** | **0.001842** |
| TF10 | 4.8836E-05 | 2.89869E-05 | **4.00E-15** | **6.38E-16** | 0.017933 | 0.013532 |
| TF11 | **2.7393E-08** | **5.51514E-08** | 0.568776 | 0.1042672 | 0.066355 | 0.030973 |
| TF12 | **1.87667E-08** | **2.89749E-08** | 19.83835 | 26.374228 | 2.79E-05 | 3.84E-05 |
| TF13 | **8.90491E-09** | **1.88063E-08** | 10.2783 | 7.42028 | 0.000309 | 0.000512 |
| TF14 | 6.9979 | 5.833242622 | **3.79E-07** | 6.32E-07 | 0.998004 | **1.26E-11** |
| TF15 | 0.001673093 | 0.003539145 | **0.001502** | **0.0012431** | 0.002358 | 0.003757 |
| TF16 | -0.622100333 | 0.396782974 | 0.006375 | 0.0105688 | **-1.03163** | **2.46E-06** |
| TF17 | 1.788405333 | 2.237631581 | 23.82013 | 0.2149425 | **0.397888** | **3.16E-06** |
| TF18 | 3.590623333 | 0.711917144 | 222.9682 | **9.96E-06** | **3.000142** | 0.000283 |
| TF19 | -2.670808 | 1.185307969 | 22.7801 | 0.0103584 | **-3.86278** | **9.61E-07** |

### *6.2. CEC-C06 2019 Benchmark Test Functions*

There are situations in the real world where getting an accurate solution is more crucial than the passing of time. Additionally, nearly anyone can perfect an algorithm and run it multiple times if they so choose. This means that clients are looking for the most effective algorithm for their situation, regardless of time. Ten test functions that were formed at the CEC-2019 conference [78], [79] and have been undergoing further Leo evaluation are part of this modern benchmark collection. The test functions, called "The 100-Digit Challenge", are supposed to be implemented in annual optimization competitions. See Table (12) in the appendix. Table (5 and 6) present the CEC-2019 functions for common and recent algorithms which are incredibly competitive and extensively applied to tackle real-world issues. Such as DA, WOA, SSA, FDO, LPB, and FOX. Furthermore, we chose these algorithms since they are all widely cited literature. They have demonstrated excellent performance both on benchmark test functions and on actual problems. The creators of these algorithms have made their implementations available to the public.

Functions CEC01 to CEC03 have different dimensions. However, the other functions have a similar dimension, which is [−100,100]. As a result, whereas functions CEC01 to CEC03 are not affected by shift and rotation, CEC04 to CEC10 are affected. All test procedures, despite this, are scalable and flexible. The developer of the CEC benchmark has defined the parameter set. The algorithms have been given the ability to run 30 times with 100 agents while performing 500 iterations of landscape search. As shown in Table (5), Leo outperforms when compared with widely cited in the literature algorithms except in the test function of CEC04. While Leo has a comparable outcome in those benchmarks, Leo and WOA have roughly similar results in benchmarks like CEC02 and CEC05. Nonetheless, as shown in Table (6), Leo surpasses FOX, a recent algorithm. Markedly, Leo performs exceptionally well in just CEC02 but for CEC06, Leo better performance rather than FDO and LPB.

Table 5. Comparing the results of Leo with DA, WOA and SSA algorithms on CEC-2019 test functions

| | Leo | | DA | | WOA | | SSA | |
|---|---|---|---|---|---|---|---|---|
| CEC | AVA | STD | AVA | STD | AVA | STD | AVA | STD |
| CEC01 | **7294147266** | **5767198154** | 5.43E+10 | 6.69E+10 | 4.11E+10 | 5.42E+10 | 6.05E+09 | 4.75E+09 |
| CEC02 | 17.47763 | 0.098108754 | 78.0368 | 87.7888 | **17.3495** | **0.0045** | 18.3434 | 0.0005 |
| CEC03 | **12.70311** | **0.000889537** | 13.7026 | 0.0007 | 13.7024 | 0 | 13.7025 | 0.0003 |
| CEC04 | 69.86527333 | 23.78089555 | 344.3561 | 414.0982 | 394.6754 | 248.5627 | **41.6936** | **22.2191** |
| CEC05 | 2.760246667 | 0.432754261 | 2.5572 | 0.3245 | 2.7342 | 0.2917 | **2.2084** | **0.1064** |
| CEC06 | **3.01982** | **0.755956506** | 9.8955 | 1.6404 | 10.7085 | 1.0325 | 6.0798 | 1.4873 |
| CEC07 | **195.5583033** | **236.5351502** | 578.9531 | 329.3983 | 490.6843 | 194.8318 | 410.3964 | 290.5562 |
| CEC08 | **5.062283333** | **0.459751941** | 6.8734 | 0.5015 | 6.909 | 0.4269 | 6.3723 | 0.5862 |
| CEC09 | **3.26147** | **0.744492954** | 6.0467 | 2.871 | 5.9371 | 1.6566 | 3.6704 | 0.2362 |
| CEC10 | **20.01238667** | **0.028550895** | 21.2604 | 0.1715 | 21.2761 | 0.1111 | 21.04 | 0.078 |

Table 6. Comparing the results of Leo with FDO, LPB and FOX algorithms on CEC-2019 test functions

| | **Leo** | | **FDO** | | **LPB** | | **FOX** | |
|---|---|---|---|---|---|---|---|---|
| CEC | AVA | STD | AVA | STD | AVA | STD | AVA | STD |
| CEC01 | 7294147266 | 5767198154 | **4585.27** | **20707.627** | 7494381364 | 8138223463 | 2.58E+04 | 22624.86 |
| CEC02 | **17.47763** | **0.098108754** | **4** | **3.22414E-09** | 17.63898 | 0.31898 | 18.3442 | 0.000529 |
| CEC03 | 12.70311 | 0.000889537 | 13.7024 | 1.649E-11 | **12.7024** | **0** | 13.7025 | 0.000449 |
| CEC04 | 69.86527333 | 23.78089555 | **34.0837** | **16.528865** | 77.90824 | 29.88519 | 1.06E+03 | 501.8163 |
| CEC05 | 2.760246667 | 0.432754261 | 2.13924 | 0.085751 | **1.18822** | **0.10945** | 6.295 | 1.27819 |
| CEC06 | **3.01982** | **0.755956506** | 12.1332 | 0.600237 | 3.73895 | 0.82305 | 5.0325 | 1.285264 |
| CEC07 | 195.5583033 | 236.5351502 | **120.4858** | **13.59369** | 145.28775 | 177.8949 | 456.3214 | 189.4313 |
| CEC08 | 5.062283333 | 0.459751941 | 6.1021 | 0.756997 | **4.88769** | **0.67942** | 5.6778 | 0.52774 |
| CEC09 | 3.26147 | 0.744492954 | **2** | **1.5916E-10** | 2.89429 | 0.23138 | 3.7959 | 0.339462 |
| CEC10 | 20.01238667 | 0.028550895 | **2.7182** | **8.8817E-16** | 20.00179 | 0.00233 | 20.9878 | 0.005376 |

### *6.3. Statistical Tests and scalability analysis*

For comparing two or more result groups, statistical tests use parametric and non-parametric tests [68]. Assumptions must be met for a two-sample t-test; thus, the Wilcoxon Rank-Sum Test is commonly referred to as the non-parametric variant [80], [81]. For each test function, the Wilcoxon Rank-Sum Test is determined to demonstrate the outcomes shown in Tables (3, 4, 5, and 6). In Tables 7 and 8, we present the results of statistical analyses. It occurs in distributions where the samples are equal or greater than 30, the assumptions are roughly equivalent or symmetric, but the spread variance and normalcy are not sufficiently satisfied.

DA results are statistically significant when compared to PSO and GA, according to research. Because DA was already evaluated against both PSO and GA in this paper [77], as well as both FDO and LPB algorithms, statistical comparisons of Leo against DA, FDO, and LPB algorithms are presented in Table (7). Meanwhile, all statistical tests (unimodal, multimodal, and composite test functions) consider the Leo results to be significant and reject the null hypothesis, except for TF6 and TF12 when DA has been conducted, because the values are greater than 0.05. Similarly, Leo’s results are positive when compared with FDO and LPB except in TF5 and TF15, which retained null hypnosis. In addition, TF17 did not demonstrate significant value with FDO. There are unusual results in all composite test functions which retain null hypnosis in DA, FDO, and LPB algorithms for each of the 30 different individual tests. Table (8) shows statistical comparisons of Leo with DA, SSA, WOA, FDO, and FOX algorithms presented by the Wilcoxon Rank-Sum Test. Leo results are highly significant and reject the null hypothesis when using DA and FOX because the values are significantly less than 0.05. DA results are statistically significant when compared to SSA, WOA, and FDO except for CEC01 with SSA, CEC05 with WOA, and CEC07 with FDO.

Table 7. $P - value$ by the Wilcoxon rank-sum test overall runs for classical benchmark test functions.

| TF | Leo VS DA | Leo VS LPB | Leo VS FDO |
|---|---|---|---|
| TF1 | 0.000031 | 0.000031 | 0.000002 |
| TF2 | 0.000002 | 0.000002 | 0.047162 |
| TF3 | 0.000002 | 0.000002 | 0.002585 |
| TF4 | 0.000031 | 0.000002 | 0.000002 |
| TF5 | 0.000148 | **0.781264** | **0.557743** |
| TF6 | **0.057096** | 0.000002 | 0.000002 |
| TF7 | 0.000002 | 0.000097 | 0.000002 |
| TF8 | 0.031603 | 0.000002 | 0.000016 |
| TF9 | 0.000002 | 0.000002 | 0.000002 |
| TF10 | 0.000002 | 0.000002 | 0.000002 |
| TF11 | 0.000002 | 0.000002 | 0.000002 |
| TF12 | **0.328571** | 0.000002 | 0.000002 |
| TF13 | 0.517048 | 0.000002 | 0.000002 |
| TF14 | 0.000013 | 0.000013 | 0.002929 |
| TF15 | 0.000359 | **0.012453** | **0.781264** |
| TF16 | 0.000001 | 0.000002 | 0.000115 |
| TF17 | 0.000001 | 0.000002 | **0.120288** |
| TF18 | 0.00015 | 0.000393 | 0.00015 |
| TF19 | 0.000002 | 0.000002 | 0.000004 |

Table 8. $P - value$ by the Wilcoxon rank-sum test overall runs for CEC-2019 test functions.

| TF | Leo VS DA | Leo VS SSA | Leo VS WOA | Leo VS FDO | Leo VS FOX |
|---|---|---|---|---|---|
| CEC01 | 0.000012 | **0.360039** | 0.038723 | 0.000002 | 0.000002 |
| CEC02 | 0.000002 | 0.000002 | 0.000002 | 0.000002 | 0.000002 |
| CEC03 | 0.000001 | 0.000001 | 0.000001 | 0.000001 | 0.000001 |
| CEC04 | 0.000012 | 0.000125 | 0.000002 | 0.000005 | 0.000002 |
| CEC05 | 0.033264 | 0.000026 | **0.688359** | 0.000004 | 0.000002 |
| CEC06 | 0.000002 | 0.000002 | 0.000002 | 0.000002 | 0.000002 |
| CEC07 | 0.000359 | 0.011079 | 0.000115 | **0.171376** | 0.000148 |
| CEC08 | 0.000002 | 0.000002 | 0.000002 | 0.000008 | 0.000082 |
| CEC09 | 0.000003 | 0.002765 | 0.000002 | 0.000002 | 0.001593 |
| CEC10 | 0.000002 | 0.000002 | 0.000002 | 0.000002 | 0.000002 |

*6.4. Quantitative Measurement Metrics*

The proposed measurement metrics are used in this short subsection to observe and analyze the anticipated Leo algorithm's performance in more detail. This experiment aims to verify convergence and forecast how the Leo algorithm could perform while handling real-world problems. The first metric measures convergence and illustrates how vaccinations promote immune responses in the body landscape. As a result, in each experiment, the benchmark functions are selected according to the unimodal benchmark functions, the multimodal test functions, and the composite benchmark functions respectively. Figure (7) depicts the agent rapidly exploring the entire region and gradually adjusting to optimality and the offspring find the best position as specified in each experiment of (FT2), (TF10), and (TF17). The experiment is shown by recording the agents' performance throughout the test from the start to the end of the test. The second experiment in each of (FT2), (TF9), and (TF17), shown in Figure (8), has Leo begin with a high fitness value and then gradually reduce it until it achieves the ideal number. When multimodal tests are conducted though, the functions are rapidly developed throughout a few iterations. The average fitness value of all Leo agents decreased noticeably throughout the iterations, according to the third test metric displayed in Figure (9) in each of (FT2), (TF9), and (TF17). According to the results, the algorithm not only enhances the global most effective agent, but it also enhances the optimal solution for all agents as well. The fourth measure records the global most optimal agent's convergence throughout each iteration. This demonstrates that the global optimized agent becomes more accurate as the number of iterations increases. Due to the focus on local search and exploitation, a dramatic shift can be seen in Figure (10) in each experiment of (FT2), (TF8), and (TF17).

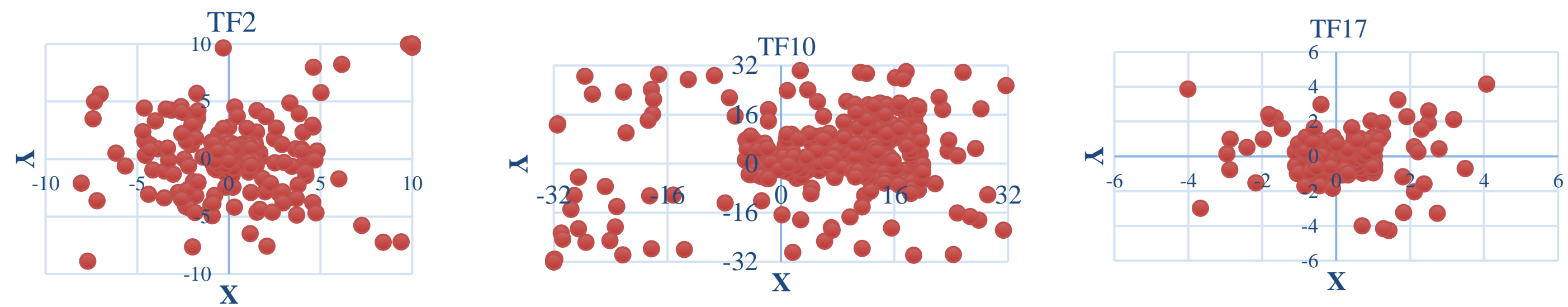

Figure 7. Search history of the Leo algorithms on unimodal, multimodal, and composite test functions

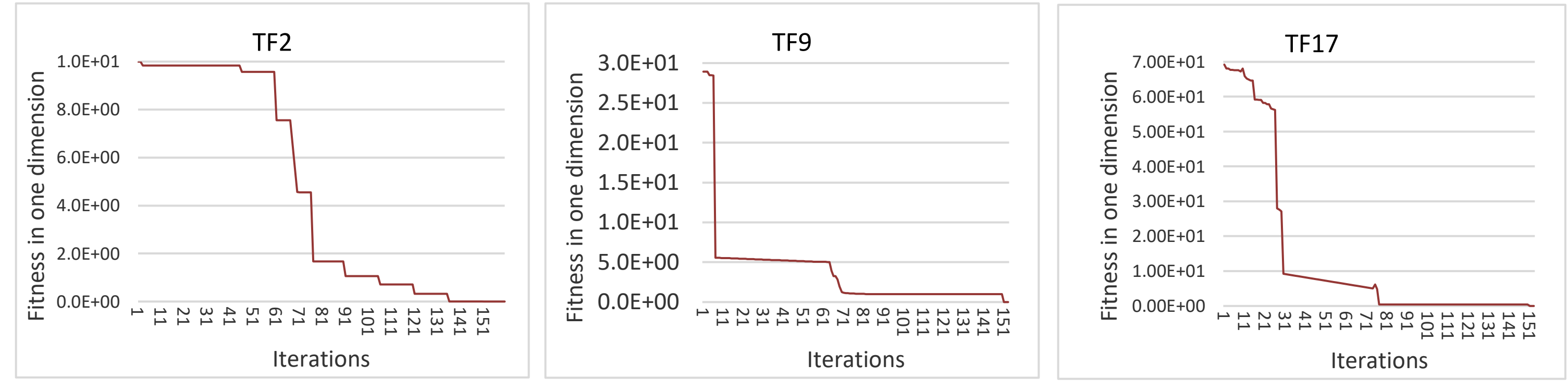

Figure 8. The trajectory of Leo's search agents on unimodal, multimodal, and composite test functions

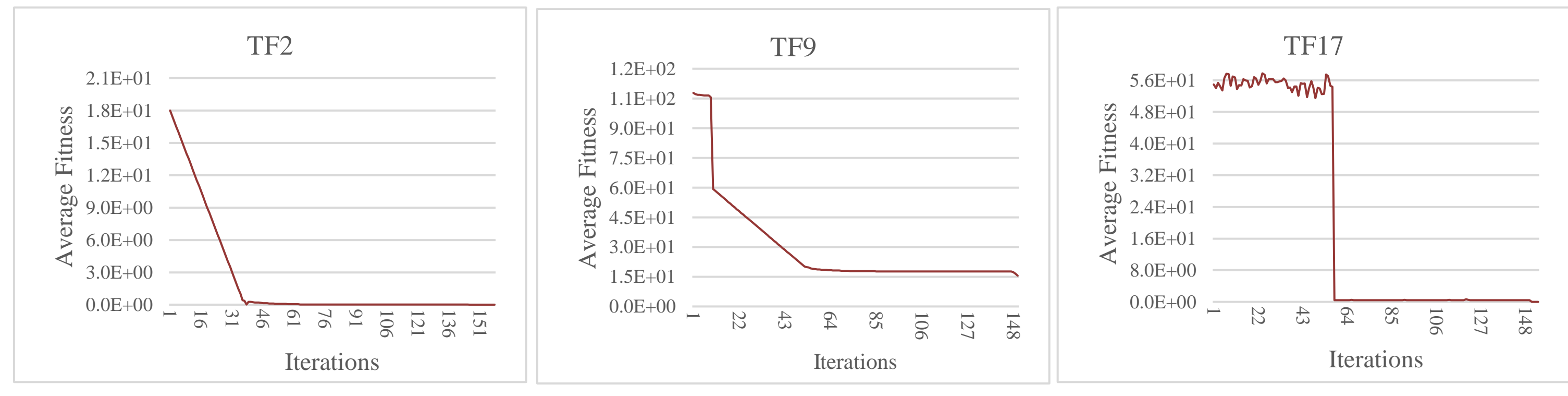

Figure 9. The average fitness of Leo's search agents on unimodal, multimodal, and composite test function

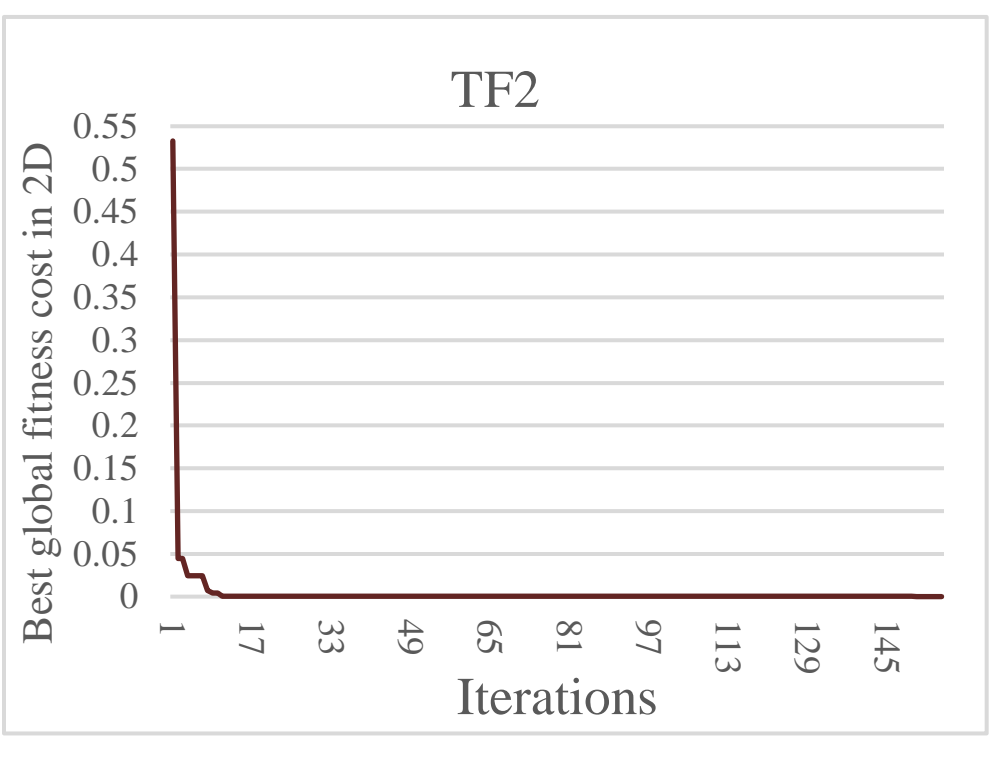

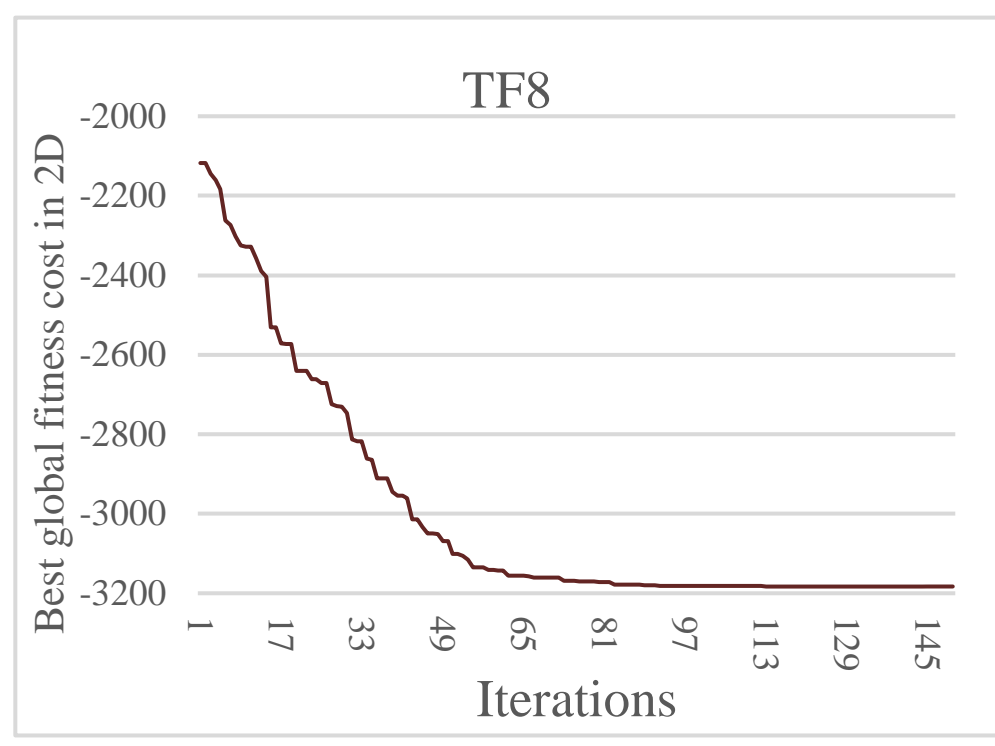

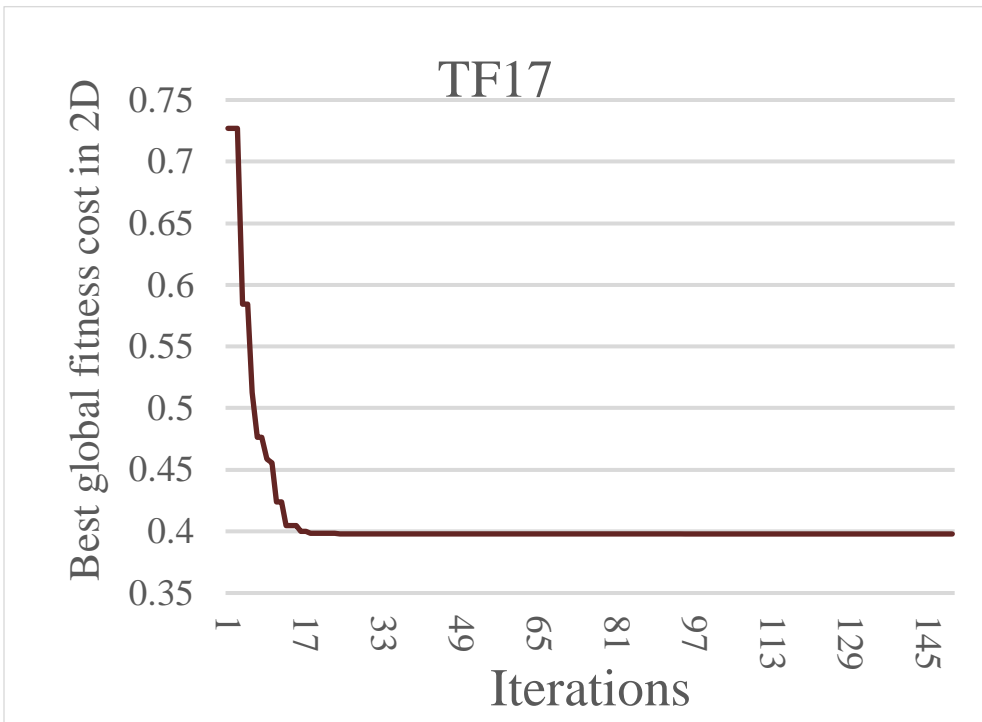

Figure 10. Convergence curve of Leo algorithms on unimodal, multi-modal, and composite test function

## 7. Real-world Application

As with any other metaheuristic algorithm, Leo can be used to overcome application-specific difficulties in the real world. Leo is used in two distinct applications in this section:

### *7.1. The Pathological IgG Fraction in the Nervous System*

The method of determination is independent of variables that may affect the individual, including sex, blood-brain barrier condition, extraction volume of CSF, and the method used to measure protein. It allows for the optimal evaluation of pathogenic IgG values in CSF when compared with other methods in the literature, particularly when it comes to statistics and biochemistry [82]. Finding the ideal solution is the aim of this problem for aspects of optimal evaluation of pathological IgG values in CSF caused to point out the fluctuation in the nervous system. Based on statistical and functional reasons, the frequency of the regression line passing through the origin is reasonable according to Eq. 12 which is enhanced from the collection of statistical regression lines [83], [84]. Most of the studies focused on establishing a correlation between serum and fluid albumin concentrations. This real application shows a correlation between serum albumin levels and IgG levels in cerebrospinal fluid and selects the optimal point by applying the Leo algorithm.
Eq. 11 [84] can be used to determine the locally generated concentration of pathological ($IgGp$) in $CSF$ after evaluating the IgG quotient for the patient's unique albumin ratio. Additionally, $STD_{(x,y)}$ is the standard deviation of the ($y$) values from the regression line between (-0.001, +0.001), and the confidence interval of the $IgG$ quotient ($y$) for a given albumin quotient ($x$) is supplied by these two variables.
This newly developed optimizer is applied to a real-world scenario, as outlined in Eq. 12, to find the IgG fraction in the nervous system and determine the optimal solution for pathological conditions in humans or animals is known as *dimmer solution*. The solution is identified using linear regression, as shown in Figure (11), within the defined range between the upper and lower bounds of (-0.001, +0.001). The regression line and the standard deviation of IgG values are used to refine the confidence interval for a more accurate evaluation to find *dimmer solutions*. Equation 11 calculates the pathological IgG concentration, while Equation 12 provides the sum of optimized values for this solution. This real-world application is summarized mathematically and medically in Figure (11).

$$IgGp = IgG(CSF) - (0.43\, Alb(Serum) - Alb(CSF) + 0.001) * IgG(Serum) \qquad \text{Eq. 11}$$

To prove that

$$IgGp = X_i \quad \text{So, IgG (IgGp)} = \text{Y}(\text{X}_\text{i}) \text{ then:}$$

$$Y\,(X_i) = \sum_{i=1}^{n} (0.41 + 0.0014\, X_i) \qquad \text{Eq. 12}$$

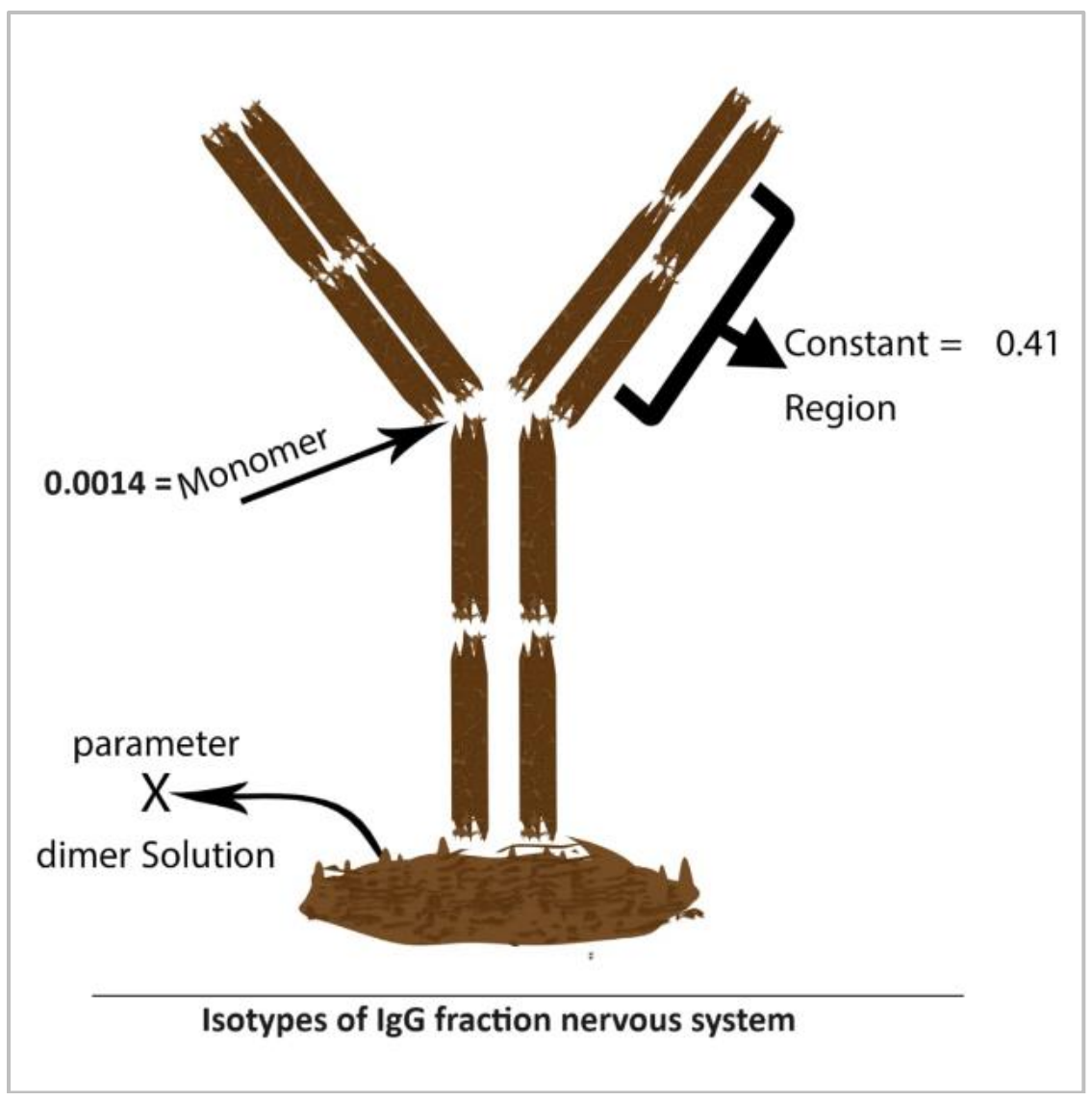

Figure 11: *Pathological $IgG$* fraction design problem

On the stained gel strip, a cutting sequence for obtaining twelve fractions has been seen. Therefore, twelve are used to calculate the number of search agents. Stippled regression lines display the $IgGp$ level in each proportion. Neutralization $(NT)$ and $CF$ are the two antibody actions that are displayed for statistical-level results [85]. With the limitations of Equation (12) in mind, this problem is optimized using the Leo algorithm. The outcome is displayed in Figure (12) and contains the global average fitness in each iteration as well as the average fitness value. Twelve search agents are employed for 150 iterations. The analysis reveals that iteration 61 of the globally optimized solution produced the optimal result, which is (5.088).

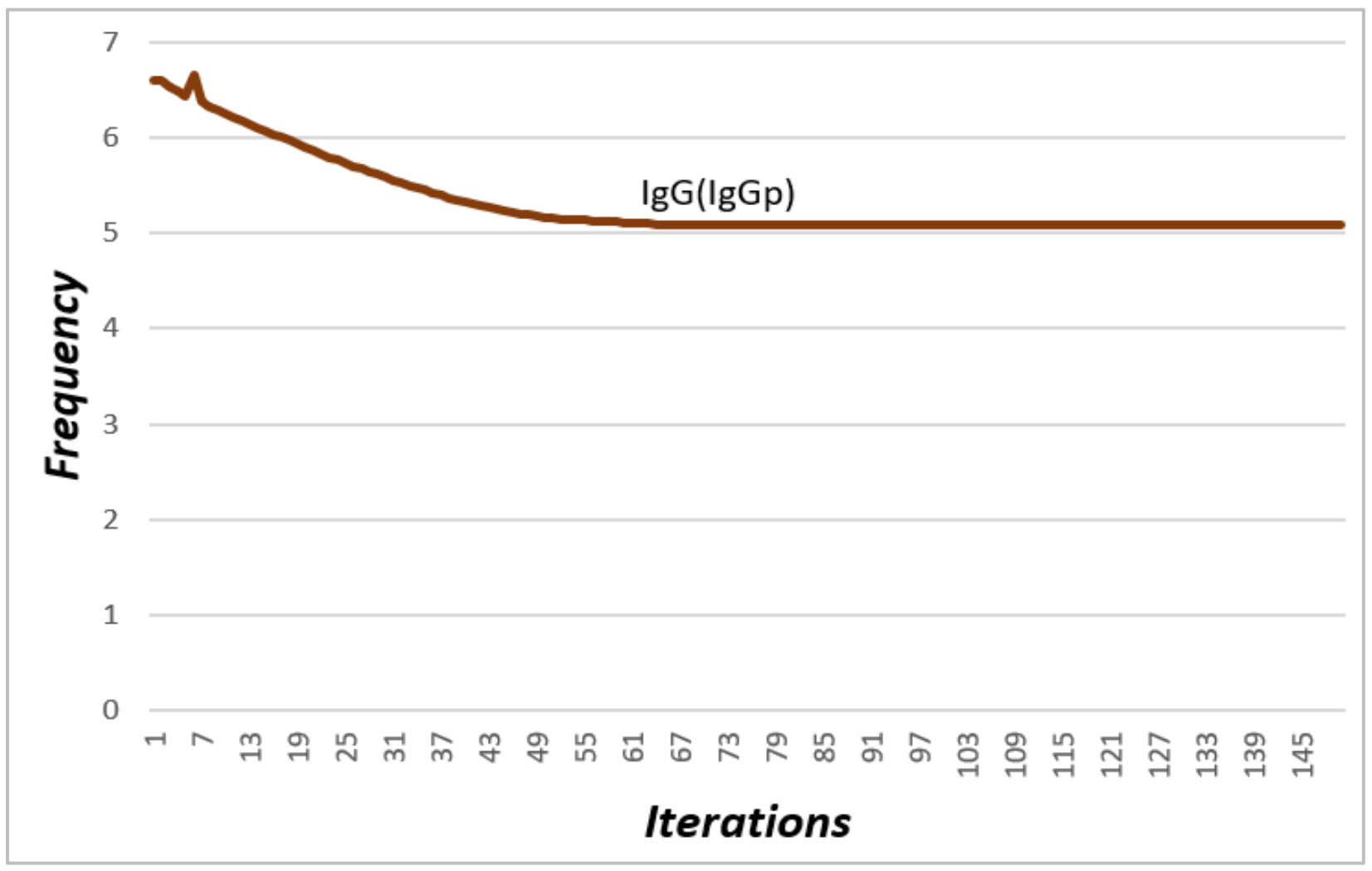

Figure 12. Global best with average fitness results from for150 Iteration with 12 search agents in (IgGp) fraction in the nervous system

*7.2. Integrated Cyber-Physical-Attack for Manufacturing System*

Generally, computational analysis can assist the security community in better comprehending the danger, analyzing the attacker's behavior during the process of a cyber-physical-attack, and as a result, being able to offer genuine responses to adversarial behaviors at various stages of an attack. Besides the paucity of research on estimating and assessing the effectiveness of defensive systems, particularly from a security standpoint, we still need to develop a suitable theoretical application model to find the global point. The use or integrate of cyber-physical-attack for manufacturing systems (CPAMS) can increase flexibility and reactivity while ensuring product quality to satisfy client demands [86]. The object-oriented Petri net-based formal model of a cyber-physical-attack manufacturing system is described from the perspective of complex systems to increase the integrity of cyber-physical systems during the dynamic simulation phase [87], [88]. This system can be verified and validated by optimizing the Leo system using some mathematical techniques and Petri net supporting tools.

Petri nets [89] are visual representations and analysis tools for distributed systems. They are updated to visualize and evaluate a wide range of applications and complex systems, including communication networks, healthcare systems, artificial intelligence, and engineering systems used in manufacturing. Besides, it is a computational mathematics tool for simulating and examining dynamic systems. A directed graph model is formed with arcs ($F$) connecting two sets of nodes, locations ($P$) and transitions ($T$). Dots inside the spots stand in for tokens (or 'marks'). There are illustrated in Figure (13) when $R1$ is explicitly defined in the net interpretation among $T$ and $P$.

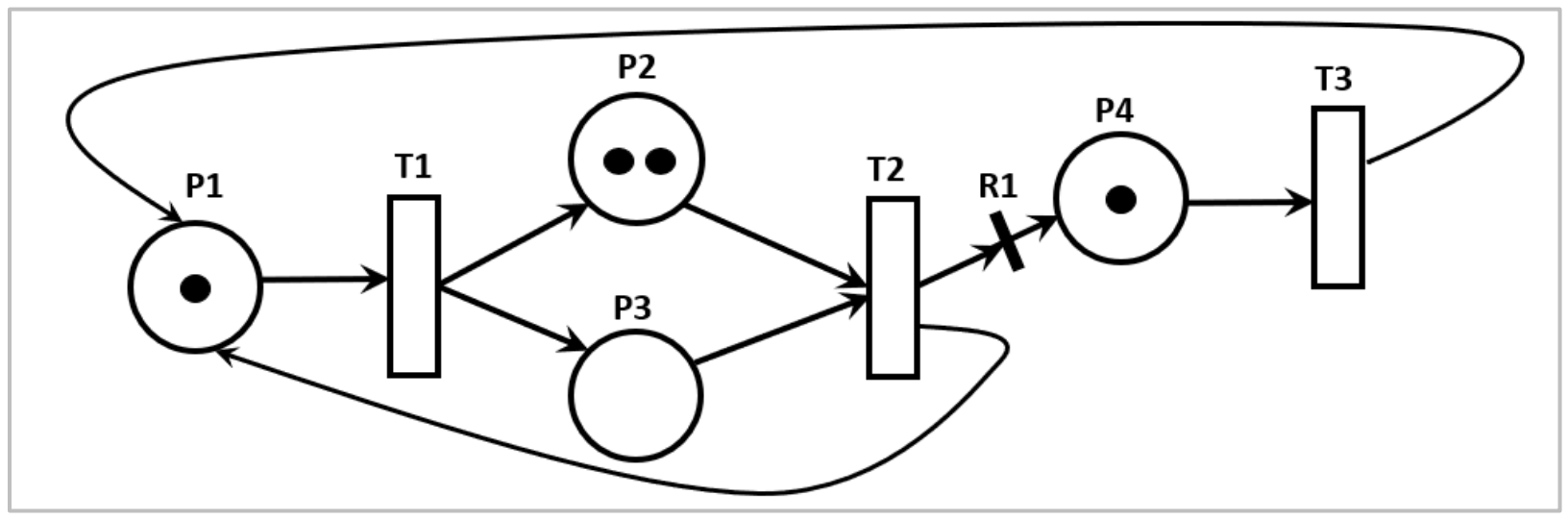

Figure 13. The network station is represented by a stochastic Petri net.

Syntax: A *Petri net* is a net of the form $PN = (N, M, W)$, which extends the elementary net so that

Given a net $N = (P, T, F)$, a configuration is a set $C$ so that $C \subseteq P$.

$M: P \rightarrow Z$ is a place multiset, where is a countable set, covers the concept of *configuration* and is normally described concerning Petri net diagrams as a *marking*.

$W: F \rightarrow Z$ is an arc multiset. The count (or weight) for each arc indicates the *multiplicity* of arcs that can be calculated.

Its transition relation can be described as a pair of $|P|$ by $|T|$ matrices:

$F^-$ , defined by $\forall s, t: F^-[p, t] = F(p, t)$

$F^+$, defined by $\forall s, t: F^+[p, t] = F(t, p)$

When, the *pre-set* of a transition *t* is the set of its *input places*: *t = {p $\epsilon$ P) | F(p,t) >0}; its *posset* is the set of its *output places*: t* = {p $\epsilon$ P) | F(p,t) >0};. Definitions of pre- and post-sets of places are analogous.

Petri net syntax states that it evaluates the system by taking into account a sizable number of CPAMS, where some nodes (such as machines, robots, sensors, and AGVs) have been infected by malicious software and spread it to susceptible nodes, which are referred to as nodes that are most vulnerable to malicious software attacks. These nodes eventually become the infected nodes. Infectious nodes can flip into recovered nodes once the harmful software has been uninstalled. The susceptible nodes, infectious nodes, and recovered nodes are symbolized, respectively, as $S$, $I$, and $R$ [90]. With the attributes of Logistic, new nodes are regarded as susceptible nodes, and their growth rate is $p$.

Control of harmful software bifurcation in CPAMS. A hybrid bifurcation law control strategy was suggested by [91] to control harmful bifurcations as indicated in Eq.13. These bifurcations result in negative behaviours and decrease the trustworthiness of CPAMS. Where $K_1$ denotes the variable parameter, and $K_2$ denotes the feedback parameter.

$$N(I(p,t)) = k_1 F(I(p,t)) + k_2(I(p,t) + I(p,t)^3) \quad \text{Eq. 13}$$

This real-world application has developed as a novel optimization tool designed to address harmful nodes in network systems, specifically during software and hardware engineering design. The primary goal is to find the optimal solution for detecting and controlling these harmful nodes between the level of [0,1], which can lead to system vulnerabilities such as bugs and risk points. The approach involves dividing key parameters between node positions and the transitions across different system layers. By doing so, the application enhances the ability to manage and mitigate potential threats effectively. This new application, known as a Cyber-Physical-Attack mitigation system for engineering networks, is thoroughly illustrated and explained in Figure (14).

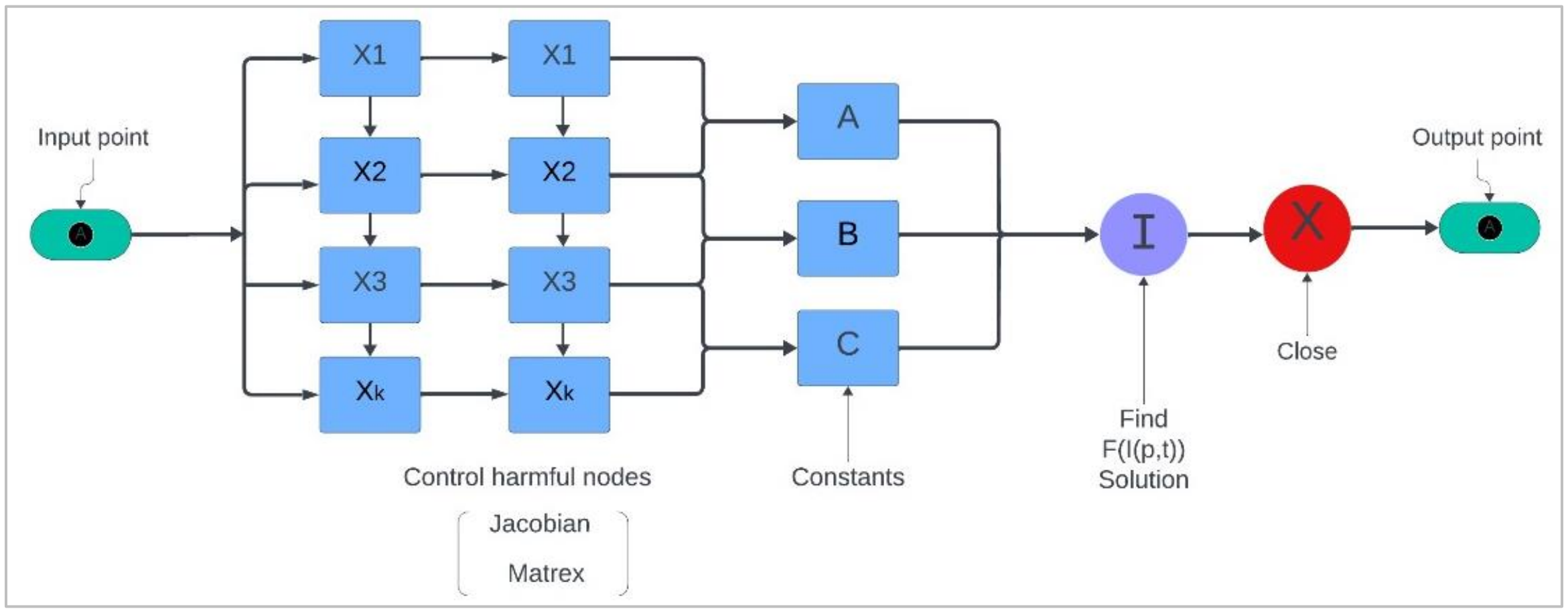

Figure 14. Design problem for a Cyber-Physical Attack mitigation system

To find the evaluation probabilistic complex system, the fitness function $F(I(p,t)$ derive from the Jacobian matrix at the equilibrium [92] to get simulation results and the optima point which controls harmful node probabilistic point and is updated by the susceptible node. Thus, the probabilistic Eq.14 helps to find the node point when equivalent to Zero.

$$F((I(p,t)) = \sum X^3 + A\sum X^2 + B\sum X + C \quad \text{Eq. 14}$$

When,

$$A = 0.0283 \left(1 + \frac{1}{d} - k_2\right)$$

$$B = \frac{0.0283 - 1.0283k_2}{d}$$

$$C = \frac{0.0013\, k_1 - 0.0283k_2}{d}$$

Suppose, set $d$ : number of nodes = 15 to 36, $k_1$ = 0 to 1, $k_2$ = 0.1 to 0.5, respectively. The Leo algorithm is used to discover the optimal approach for updating infection nodes in this realistic situation. The result includes both the average fitness value and the global average fitness for each iteration. 300 iterations are performed using 10 search agents. The study shows that the globally optimized solution's iteration 209 generated the most successful outcome, which is (.072028). Figure (15) illustrates the process.

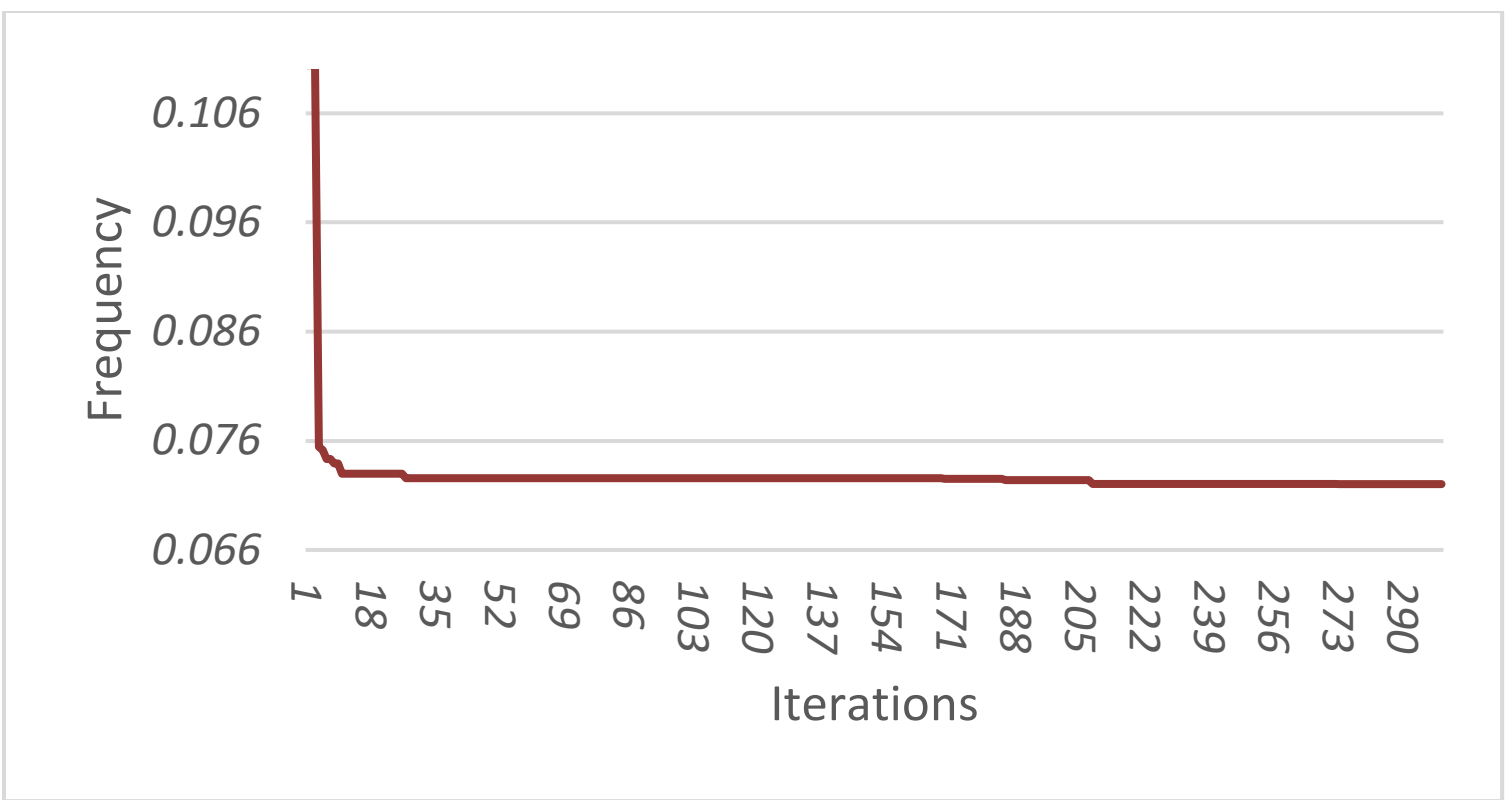

Figure 15. Fitness results in Leo process for 300 Iteration with 10 search agents depend on the Jacobian matrix for cyber-physical-attack in the manufacturing system.

### *7.3. Leo Real-world Application Comparison and Discussion*

The efficiency of these real-world optimization algorithms has been assessed as a novel contribution to this study, focusing on their ability to generate new solutions. The comparison demonstrates how each algorithm influences solution discovery, with both applications showing significant potential in achieving new applications. The dynamic nature of these algorithms emphasizes their convergence behavior, which is critical for solving complex problems. Additionally, the results indicate that both the real-time performance and convergence-stopping criteria are successfully met in these applications, highlighting their high effectiveness in problem-solving. To illustrate this effect, we compare the Leo algorithm, as developed in this study, with the Whale Optimization Algorithm (WOA) and the Dragonfly Algorithm (DA) across both real-world applications. Figure 16-A presents a comparison of these three algorithms for the Pathological IgG Fraction System application. The results indicate that the Leo algorithm initially starts with a smaller value but successfully finds the minimum more effectively. In contrast, the DA reaches a slightly smaller optimal solution and converges more easily. Meanwhile, the Leo algorithm shows a gradual minimization, demonstrating its effectiveness in finding solutions within the selected search space. Additionally, the study compares the performance of the algorithms for the CPAMS Real-World Application in cyber-physical attack security, as demonstrated in Figure 16-B. The assessment shows that the starting points of the Leo and DA algorithms are fairly similar. However, over time, the DA algorithm converges more rapidly, achieving a solution sooner and with fewer adjustments. In contrast, the Leo algorithm exhibits a more gradual convergence with continuous but minimal modifications. The WOA algorithm, starting from a higher initial point, ultimately finds a smaller convergence solution compared to Leo, but takes longer to achieve this result.

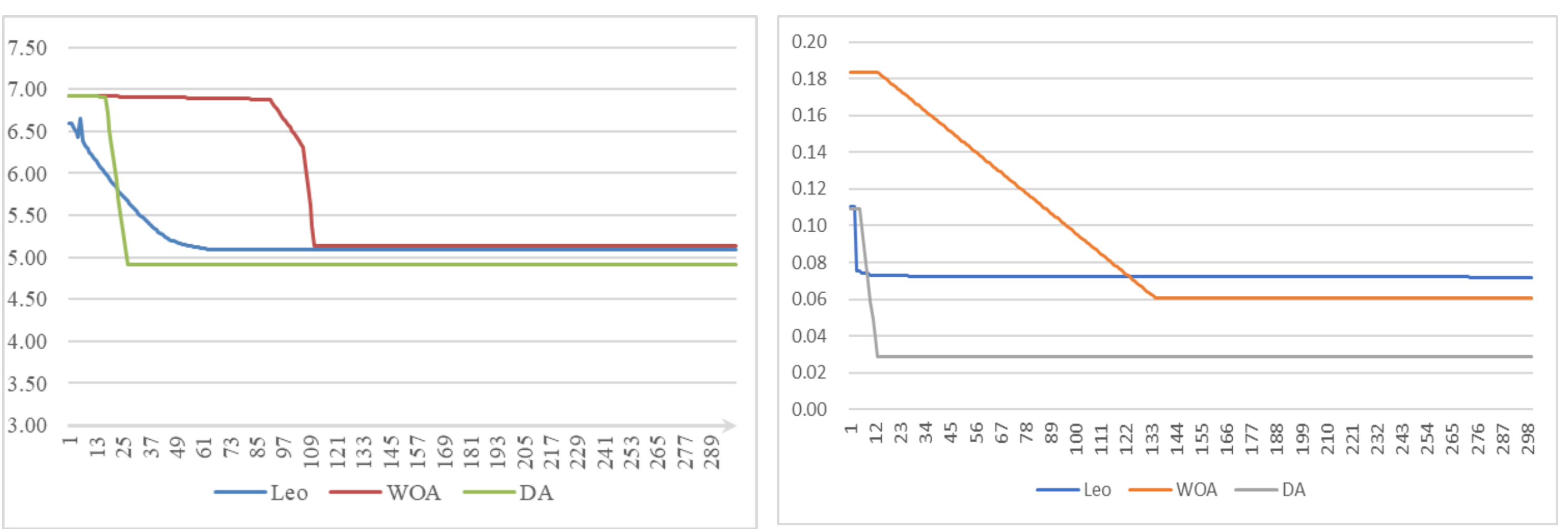

*A-* *Pathological IgG* fraction Real-world Application *B-* CPAMS Real-world Application

Figure 16. Compare the real-world efficiency of the Leo, WOA, and DA algorithms in finding the best solutions.

## 8. Conclusion

To achieve exceptional precision of immunizations utilizing the albumin quotient of human blood, this research developed an innovative single-objective metaheuristic algorithm based on the process of transferring genetic chromosomes. This algorithm was inspired by the genetic algorithm. Exploitation and exploration—the two key stages of metaheuristic algorithms—were described. The previous stage was modeled after the division of albumin into various groups based on blood serum and their gene parents. Nevertheless, the exploration phase was designed to simulate how many effective vaccinations raise the immune system. Crossover and mutation parameters were the factors employed in the Leo, which are used in the initial stages of the algorithm to divide the population into multiple groups and introduce success-based population distribution. Using crossover and mutation techniques, the Leo was auto-adaptive when Lagrange orientation and Lagrange multiplier stationary point navigation were introduced.

By evaluating numerous common test functions and real-world applications, we discovered that, typically, the quantity of search agents was somehow related to performance. Because the algorithm depends on the Lagrange stationary point during gene crossover for a significant portion of its searching mechanism, it is known as the Lagrange Elementary for Optimization, and using a small number of agents (less than seven as the median for all benchmarks) would significantly decrease the accuracy of the algorithm while using a large number of search representatives would improve the accuracy while costing more genes and updating of offspring. Based on the results of the study, it can be concluded that the proposed work outperforms most algorithms in the field. So far, Leo may have difficulties with growing problem sizes for inspired optimization. Therefore, using it as an optimization method is recommended for researchers in various domains.

Moreover, the initialization, exploration, and exploitation stages of Leo rely on the randomization approach. To evaluate the performance of Leo, 19 single-objective benchmark testing functions were used. Three separate subgroups were created for the benchmark testing functions (unimodal, multimodal, and composite test functions). Leo also ran tests on 10 current CEC-C06 benchmarks. Leo outperformed the rival algorithms in the majority of situations when it was compared to two well-known algorithms (PSO and GA), three modern algorithms (DA, WOA, and SSA), and three recent algorithms (FDO, LPB, and FOX). The Wilcoxon rank-sum test was used to compare the test results to demonstrate their statistical significance. To prove and verify the performance, the method can handle real-world applications, Leo was also practically applied to two real-world applications.

In future work, it will be adapted, implemented and tested on Leo for both multi-objective and binary objective optimization problems. Furthermore, future research could incorporate evolutionary operators into Leo and hybridize them with other algorithms. Moreover, a new Lagrange mutation standard will replace these parameters from Leo in the future. Finally, the proposed technique can be used to optimize a variety of problems and compare the results to those of other heuristic techniques.

**Appendix:** This work applies single-objective test problems. The mathematical representations of the traditional benchmark functions used in this preparation are shown in Tables 9, 10, and 11. The CEC-C06 2019 benchmark functions are shown in Table 12.

Table 9. Unimodal benchmark functions [76]

| Functions | Dimension | Range | Shift position | $f_{min}$ |
|---|---|---|---|---|
| $TF1(x) = \sum_{i=1}^{n} x_i^2$ | 10 | [-100, 100] | [-30, -30, … -30] | 0 |
| $TF2(x) = \sum_{i=1}^{n} \lvert x_i \rvert + \prod_{i=1}^{n} \lvert x_i \rvert$ | 10 | [-10,10] | [-3, -3, … -3] | 0 |
| $TF3(x) = \sum_{i=1}^{n} \left( \sum_{j-1}^{i} x_j \right)^2$ | 10 | [-100, 100] | [-30, -30, … -30] | 0 |

| $\boldsymbol{TF4(x)} = \max_{i}\{\lvert x\rvert, 1 \le i \le n\}$ | 10 | [-100, 100] | [-30, -30, … -30] | 0 |
|---|---|---|---|---|
| $\boldsymbol{TF5(x)} = \sum_{i=1}^{n-1}[100(x_{i+1} - x_1^2)^2 + (x_i - 1)^2]$ | 10 | [-30,30] | [-15, -15, … -15] | 0 |
| $\boldsymbol{TF6(x)} = \sum_{i=1}^{n}([x_i + 0.5])^2$ | 10 | [-100, 100] | [-750, … -750] | 0 |
| $\boldsymbol{TF7(x)} = \sum_{i=1}^{n} i x_i^4 + \text{random}[0, 1]$ | 10 | [-1.28,1.28] | [-0.25, …-0.25] | 0 |

Table 10. Multimodal benchmark functions (10 dimensional) [76]

| **Functions** | **Range** | **Shift position** | $\boldsymbol{f_{min}}$ |
|---|---|---|---|
| $\boldsymbol{TF8(x)} = \sum_{i=1}^{n} -x_i \sin\left(\sqrt{\lvert x_i\rvert}\right)$ | [-500, 500] | [-300, … -300] | -418.9829 |
| $\boldsymbol{TF9(x)} = \sum_{i=1}^{n}\left[x_i^2 - 10\cos(2\pi x_i) + 10\right]$ | [-5.12,5.12] | [-2, -2, …-2] | 0 |
| $\boldsymbol{TF10(x)} = \boldsymbol{-20exp\left(-0.2\sqrt{\sum_{i=1}^{n} x_i^2}\right) - exp\left(\frac{1}{n}\sum_{i=1}^{n} cos(2\pi x_i)\right) + 20 + e}$ | [-32, 32] | | 0 |
| $\boldsymbol{TF11(x)} = \frac{1}{4000}\sum_{i=1}^{n} x_i^2 - \prod_{i=1}^{n}\cos\left(\frac{x_i}{\sqrt{i}}\right) + 1$ | [-600, 600] | [-400, … -400] | 0 |
| $\boldsymbol{TF12(x)} = \frac{\pi}{n}\{10\sin(\pi y_1) + \sum_{i=1}^{n-1}(y_i - 1)^2[1 + 10\ \sin^2(\pi y_{i+1})] + (y_n - 1)^2\} + \sum_{i=1}^{n} u(x_i, 10, 100, 4).$<br>$y_i = 1 + \frac{x+1}{4}.\qquad u(x_i, a, k, m) = \begin{cases} k(x_i - a)^m\ x_i > a \\ 0\ -a < x_i < a \\ k(-x_i - a)^m\ x_i < -a \end{cases}$ | [-50,50] | [-30, 30, … 30] | 0 |
| $\text{TF13(x)} = 0.1\{\sin^2(3\pi x1) + \sum_{i=1}^{n}(x_i - 1)^2[1 + \sin^2(3\pi x_i + 1)] + (x_n - 1)^2[1 + \sin^2(2\pi x_n)]\} + \sum_{i=1}^{n} u(x_i, 5,100,4).$ | [-50,50] | [-100, … -100] | 0 |

Table 11. Composite benchmark functions [76]

| **Functions** | **Dimension** | **Range** | $\boldsymbol{f_{min}}$ |
|---|---|---|---|
| **TF14 (CF1)** $f1, f2, f3 \ldots f10$ = Sphere function $\delta1, \delta2, \delta3 \ldots \delta10$ [1,1,1, … .1] $\lambda1, \lambda2, \lambda3 \ldots \lambda10 = \left[\frac{5}{100}, \frac{5}{100,}, \frac{5}{100}, \ldots \frac{5}{100}\right]$ | 10 | [-5, 5] | 0 |
| **TF15 (CF2)** $f1, f2, f3 \ldots f10$ Griewank's function $\delta1, \delta2, \delta3 \ldots \delta10$, [1,1,1, … .1] $\lambda1, \lambda2, \lambda3 \ldots \lambda10 = \left[\frac{5}{100}, \frac{5}{100,}, \frac{5}{100}, \ldots \frac{5}{100}\right]$ | 10 | [-5, 5] | 0 |
| **TF16 (CF3)** $f1, f2, f3 \ldots f10$ Griewank's function $\delta1, \delta2, \delta3 \ldots \delta10$, [1,1,1, … .1] $\lambda1, \lambda2, \lambda3 \ldots \lambda10$ = [1,1,1, … .1] | 10 | [-5, 5] | 0 |
| **TF17 (CF4)** $f1, f2$ = Ackley's function, $f3, f4$ = Rastrigin's function, $f5, f6$ = Weierstrass functio, $f7, f8$ = Griewank's function, $f9, f10$ = Sphere function, $\delta1, \delta2, \delta3 \ldots \delta10$ = [1,1,1, … .1] $\lambda1, \lambda2, \lambda3 \ldots = \left[\frac{5}{32}, \frac{5}{32,}, 1, 1, \frac{5}{0.5}, \frac{5}{0.5}, \frac{5}{100}, \frac{5}{100}, \frac{5}{100}, \frac{5}{100}\right]$ | 10 | [-5, 5] | 0 |
| **TF18 (CF5)** $f1, f2$ = Rastrigin's function, $f3, f4$ = Weierstrass function, $f5, f6$ = Griewank's function, $f7, f8$ = Ackley's function $f9, f10$ = Sphere function, $\delta1, \delta2, \delta3 \ldots \delta10$ = [1,1,1, … .1] $\lambda1, \lambda2, \lambda3 \ldots \lambda10 = \left[\frac{1}{5}, \frac{1}{5,}, \frac{5}{0.5}, \frac{5}{0.5}, \frac{5}{100}, \frac{5}{100}, \frac{5}{32}, \frac{5}{32}, \frac{5}{100}, \frac{5}{100}\right]$ | 10 | [-5, 5] | 0 |
| **TF19 (CF6)** $f1, f2 =$ Rastrigin's function, $f3, f4 =$ Weierstrass function, $f5, f6 =$ Griewank's function, $f7, f8 =$ Ackley's function, $f9, f10$Sphere function, $\delta1, \delta2, \delta3 \ldots \delta10$ [0.1,0.2,0.3, 0.4,0.5,0.6,0.7,0.8,0.9,1], $\lambda1, \lambda2, \lambda3$ .. $\left[0.1 * \frac{1}{5}, 0.2 * \frac{1}{5}, 0.3 * \frac{5}{0.5}, 0.4 * \frac{5}{0.5}, 0.5 * \frac{5}{100}, 0.6 * \frac{5}{100}, 0.7 * \frac{5}{32}, 0.8 * \frac{5}{32}, 0.9 * \frac{5}{100}, 1 * 5/100\right]$ | 10 | [-5, 5] | 0 |

Table 12. CEC-2019 benchmarks "the 100-digit challenge" [79]

| No. | Functions | Dimension | Range | $f_{min}$ |
|---|---|---|---|---|
| 1 | STORN'S CHEBYSHEV POLYNOMIAL FITTING PROBLEM | 9 | [-8192, 8192] | 1 |
| 2 | INVERSE HILBERT MATRIX PROBLEM | 16 | [-16384, 16384] | 1 |
| 3 | LENNARD-JONES MINIMUM ENERGY CLUSTER | 18 | [-4,4] | 1 |
| 4 | RASTRIGIN'S FUNCTION | 10 | [-100, 100] | 1 |
| 5 | GRIEWANGK'S FUNCTION | 10 | [-100, 100] | 1 |
| 6 | WEIERSTRASS FUNCTION | 10 | [-100, 100] | 1 |
| 7 | MODIFIED SCHWEFEL'S FUNCTION | 10 | [-100, 100] | 1 |
| 8 | EXPANDED SCHAFFER'S F6 FUNCTION | 10 | [-100, 100] | 1 |
| 9 | HAPPY CAT FUNCTION | 10 | [-100, 100] | 1 |
| 10 | ACKLEY FUNCTION | 10 | [-100, 100] | 1 |

**Compliance with ethical standards:**

**Conflict of interest:** The authors declare no conflict of interest to any party.

**Funding:** The research received no funds.

**Informed Consent Statement:** Not applicable.

**Data Availability:** Data can be shared upon request from the corresponding author.

**Acknowledgment:** None.